\documentclass[journal]{IEEEtran}
\usepackage{cite}

\newcommand{\revise}[1]{\textcolor{black}{#1}}

\ifCLASSINFOpdf
  \usepackage[pdftex]{graphicx}
  \DeclareGraphicsExtensions{.pdf,.jpeg,.png}
\else
  \usepackage[dvips]{graphicx}
  \DeclareGraphicsExtensions{.eps}
\fi
\usepackage{amsmath}
\usepackage{algorithmic}

\usepackage{array}

\ifCLASSOPTIONcompsoc
  \usepackage[caption=false,font=normalsize,labelfont=sf,textfont=sf]{subfig}
\else
  \usepackage[caption=false,font=footnotesize]{subfig}
\fi
\usepackage{url}

\usepackage{caption}

\usepackage{multirow}

\usepackage{amsfonts, amssymb}
\usepackage{amsthm}

\newtheorem{prop}{Proposition}
\usepackage{algorithm}

\usepackage{color}

\begin{document}
%
\title{Semantic Cluster Unary Loss for Efficient Deep Hashing}

\author{\IEEEauthorblockN{Shifeng Zhang, Jianmin Li, and Bo Zhang}
\thanks{Manuscript received May 14, 2018; revised October 30, 2018; accepted December 28, 2018. This work is supported by the National Key Research and Development Program of China (under Grant 2017YFA0700904), and the National Natural Science Foundation of China (Nos. 61620106010 and 61571261). The associate editor coordinating the review of this manuscript and approving it for publication was Dr. Jocelyn Chanussot. {\em (Corresponding author: Jianmin Li)}.}
\thanks{ Shifeng Zhang, Jianmin Li and Bo Zhang are with State Key Lab of Intelligent Technology and Systems, Beijing National Research Center for Information Science and Technology,
Department of Computer Science and Technology, Tsinghua University, Beijing, 100084, China. (E-mail:  \{zhangsf15@mails,lijianmin@mail,dcszb@mail\}.tsinghua.edu.cn)}
%
\thanks{This paper has supplementary downloadable material available at https://ieeexplore.ieee.org/document/8607035, provided by the author. The material includes the derivations for some equations and the proof of the proposition. Contact zhangsf15@mails.tsinghua.edu.cn for further questions about this work.}
\thanks{Digital Object Identifier 10.1109/TIP.2019.2891967}
}


%

\markboth{IEEE TRANSACTIONS ON IMAGE PROCESSING}%
{ZHANG \MakeLowercase{\textit{et al.}}: SEMANTIC CLUSTER UNARY LOSS FOR EFFICIENT DEEP HASHING}
%



\maketitle

\begin{abstract}
Hashing method maps similar data to binary hashcodes with smaller hamming distance, which has received broad attention due to its low storage cost and fast retrieval speed. With the rapid development of deep learning, deep hashing methods have achieved promising results in efficient information retrieval. Most existing deep hashing methods adopt pairwise or triplet losses to deal with similarities underlying the data, but their training are difficult and less efficient because $O(n^2)$ data pairs and $O(n^3)$ triplets are involved. To address these issues, we propose a novel deep hashing algorithm with unary loss which can be trained very efficiently. First of all, we introduce a Unary Upper Bound of the traditional triplet loss, thus reducing the complexity to $O(n)$ and bridging the classification-based unary loss and the triplet loss. Second, we propose a novel Semantic Cluster Deep Hashing (SCDH) algorithm by introducing a modified Unary Upper Bound loss, named Semantic Cluster Unary Loss (SCUL). The resultant hashcodes form several compact clusters, which means hashcodes in the same cluster have similar semantic information. We also demonstrate that the proposed SCDH is easy to be extended to semi-supervised settings by incorporating the state-of-the-art semi-supervised learning algorithms. Experiments on large-scale datasets show that the proposed method is superior to state-of-the-art hashing algorithms.


\end{abstract}


\section{Introduction}

\IEEEPARstart{D}{uring} the past few years, hashing has become a popular tool in solving large-scale vision and machine learning problems~\cite{gionis1999similarity,Shen_2015_CVPR,xia2014supervised,liu2016deep,li2011hashing,liu2012supervised,gu2016supervised}. Hashing techniques encode various types of high-dimensional data into compact hashcodes, so that similar data are mapped to hashcodes with smaller Hamming distance. With the compact binary codes, we are able to compress data into small storage space, and conduct the efficient nearest neighbor search on large-scale datasets.

The hashing techniques are categorized into {\em data-independent} methods and {\em data-dependent} methods. {\em Data-independent} methods like Locality-Sensitive Hashing (LSH) ~\cite{gionis1999similarity,charikar2002similarity,ji2013min,datar2004locality,li2010b,ji2015angular} have theoretical guarantees that similar data have higher probability to be mapped into the same hashcode, but they need relatively long codes to achieve such high precision. {\em Data-dependent} learning-to-hash methods aim at learning hash functions with training data. A number of methods are proposed in the literature, which can be summarizes as: {\em unsupervised hashing}~\cite{weiss2009spectral,gong2013iterative,liu2014discrete,kong2012isotropic,liu2017hash,hao2017unsupervised}, {\em supervised hashing}~\cite{kulis2009learning,liu2011hashing,Shen_2015_CVPR,norouzi2011minimal,lin2014fast,zhang2017scalable,lai2016instance,liu2017sequential,cakir2017mihash} and {\em semi-supervised hashing}~\cite{wang2012semi}. Experiments convey that hashcodes learned by (semi-)supervised hashing methods contain more semantic information than those learned by the unsupervised ones.

Recently, with the rapid development of deep learning~\cite{krizhevsky2012imagenet,simonyan2014very,ren2015faster}, deep hashing methods have been proposed to learn hashcodes as well as deep networks simultaneously. The codes generated by the deep networks contain much better semantic information~\cite{xia2014supervised,lai2015simultaneous,li2015feature,liu2016deep,zhu2016deep,zhuang2016fast,zhang2017scalable,li2017deep,guo2016hash,zhang2015bit,lu2017deep}, and extensive experiments convey that deep hashing methods achieve superior performance over traditional methods in a variety of retrieval tasks.

Despite the advantages of the deep hashing methods, most of them use pairwise or triplet similarities to learn hash functions to ensure that similar data can be mapped to similar hashcodes. But there are $O(n^2)$ data pairs or $O(n^3)$ data triplets where $n$ is the number of the training instances, which are too large for large scale dataset. To overcome this issue, most hashing methods~\cite{liu2016deep,zhu2016deep,li2015feature,dai2016binary} just consider generating data pairs/triplets within a mini-batch, but this approach is only able to cover limited data pairs/triplets and is hard to converge; what's worse, similar data pairs are scarce within a mini-batch if the number of the labels is too large. Some approaches like CNNBH~\cite{guo2016hash} directly regard the intermediate layer of a classification model as the hash layer to reduce the complexity to $O(n)$ and achieve good retrieval results. However, it lies in the assumption that the learned binary codes are good for linear classification, which makes is possible that the semantic gap is involved within the similar hashcodes. Efficient deep hash learning methods with low complexity are expected to be discovered.

It is clear that hashing is a special case of metric learning, which aims at learning a certain similarity function. Metric learning is widely used in many areas such as face recognition~\cite{schroff2015facenet,parkhi2015deep,sun2014deep}, (fine-grained) image retrieval~\cite{wan2014deep,movshovitz2017no} and so on. Recent approaches like CenterLoss~\cite{wen2016discriminative} and L-Softmax~\cite{liu2017sphereface} propose modified classification-based unary losses to learn a good metric, in which the intra-class distances are minimized and inter-class distances are far apart. Empirical experiments on metric learning problems such as face verification show promising results. It conveys that a good metric can be learned by optimizing a carefully designed classification-based unary loss, which motivates us to utilize it for efficient hash learning.

Moreover, in practical applications, the size of the database for retrieval is dramatically increasing to provide desired retrieval results. However, labeling all the database data is difficult and only part of the labeled data can be obtained. For generating efficient codes, deep semi-supervised hashing~\cite{yan2017semi,zhang2017ssdh} has been proposed in which the hash function is trained with the labeled data as well as abundant unlabeled data in the database. These methods construct graphs for unlabeled data, but graph based methods are not working well for complex dataset. Recent perturbation based deep semi-supervised learning(SSL) algorithm such as Temporal Ensembling~\cite{laine2016temporal} and Mean Teacher~\cite{tarvainen2017mean} has witnessed great success, but combining them with pairwise/triplet losses for hashing is difficult. By incorporating a carefully designed classification-based unary loss for hash learning with the state-of-the-art SSL algorithm, it is expected to achieve efficient semi-supervised hashing(or metric learning).

\subsection{Our Proposal}
In this paper, we propose a novel (semi-)supervised hashing algorithm with high training efficiency, in which a novel classification-based unary loss is introduced. First of all, we introduce a Unary Upper Bound of the traditional triplet loss, the latter being widely used in the hash learning scheme. The Unary Upper Bound bridges the triplet loss and the classification-based unary loss (like hinge loss, softmax). It shows that each semantic label corresponds to a certain cluster, and different clusters corresponding to different labels should be far apart. Furthermore, minimizing the Unary Upper Bound makes the intra-class distances go smaller and the inter-class distances go larger, thus the traditional triplet loss can be minimized. Second, we propose a novel supervised hashing algorithm in which we introduce a modified Unary Upper Bound loss called Semantic Cluster Unary Loss (SCUL). The complexity of SCUL is just $O(n)$, making the training procedure more efficient. Third, we introduce a novel semi-supervised hashing algorithm by incorporating the SCUL and the state-of-the-art Mean Teacher({\em MT})~\cite{tarvainen2017mean} algorithm, where the softmax loss is replaced by the SCUL. We name the proposed algorithm as {\em Semantic Cluster Deep Hashing (SCDH)} algorithm and name the semi-supervised extension as {\em MT-SCDH}.

Our main contributions are summarized as follows: 
\begin{enumerate}
\item We introduce the Unary Upper Bound of the triplet loss, thus bridging the classification-based unary loss and the triplet loss.
\item We propose a novel and efficient deep supervised hashing algorithm, which is trained with Semantic Cluster Unary Loss(SCUL), a modified Unary Upper Bound. The complexity of the SCUL is just $O(n)$ and the algorithm can be trained efficiently.
\item We propose a novel semi-supervised hashing algorithm by incorporating the Unary Upper Bound with Mean Teacher, the state-of-the-art SSL algorithm.
\item Extensive experimental results on several (semi-)supervised hashing datasets show its superiority over the state-of-the-art hashing methods.
\end{enumerate}

The rest of the paper is organized as follows. Section \ref{sec:related} presents the related work on deep hashing and semi-supervised learning. Section \ref{sec:uub} proposes the Unary Upper Bound of the triplet loss and introduces a modified form called Semantic Cluster Unary Loss (SCUL). Section \ref{sec:scdh} introduces the novel Semantic Cluster Deep Hashing (SCDH) algorithm in which the SCUL is applied, and we propose a novel extension for semi-supervised hashing in Section \ref{sec:ssh}. Experiments are shown in Section \ref{sec:experiments}, and the conclusions are summarized in Section \ref{sec:conclusion}. The codes for proposed algorithms have been released at \texttt{https://github.com/zsffq999/SCDH}.

\section{Related Work}
\label{sec:related}

\subsubsection{Deep Supervised Hashing} Recently, deep convolutional neural network (CNN) have received great success in image classification~\cite{krizhevsky2012imagenet,simonyan2014very,he2015deep}, object detection~\cite{ren2015faster} and so on. Deep hashing methods simultaneously learn hash functions as well as the network, and the hashcodes are generated directly from the deep neural network. One of the difficulties in hash learning is that the discrete constraints are involved. For ease of back-propagation, some methods remove the discrete constraints and add tanh nonlinearity~\cite{lai2015simultaneous,dai2016binary,cakir2017mihash}, or add some quantization penalty to reduce the gap between the real-valued vectors and the hashcodes~\cite{zhu2016deep,liu2016deep,li2015feature}. Some methods introduce discrete hashing methods to generate codes without relaxations~\cite{zhang2017scalable,li2017deep}. Both relaxation and discrete methods succeed in dealing with discrete constraints to some extent.

It should be noticed that different hashing methods use different type of losses for optimization. DHN~\cite{zhu2016deep}, DSH~\cite{liu2016deep}, DPSH~\cite{li2015feature} and DH~\cite{lu2017deep} use pairwise loss, which are optimized so that the hamming distance of codes with similar semantic information should be small, and vice versa. NINH~\cite{lai2015simultaneous}, BOH~\cite{dai2016binary} and DRSCH~\cite{zhang2015bit} propose triplet loss, in which the hamming distances of similar codes should be smaller than dissimilar ones by a margin. However, in these methods, at least $O(n^2)$ data pairs and $O(n^3)$ data triplets should be considered, which are too large for large-scale datasets. DISH~\cite{zhang2017scalable} overcomes this issue by performing the matrix factorization of the similarity matrix, but it is not able to be trained by direct back-propagation. CNNBH~\cite{guo2016hash} directly uses the activations of the intermediate layer as hashcodes, and the networks are trained directly from the traditional softmax loss to reduce the complexity to $O(n)$, but it assumes that the learned binary codes should be good for classification, lacking the guarantees that similar hashcodes correspond to data with similar semantic information.

\subsubsection{Metric Learning} Metric learning aims at learning a certain similarity function. Hashing is a special case of metric learning in that the similarity is defined by the Hamming distance, thus pairwise and triplet losses are also introduced to the metric learning scheme~\cite{sun2014deep,schroff2015facenet,parkhi2015deep}, but they also suffer from high complexity and low convergence. Some recent works introduce classification-based unary loss for optimization like CenterLoss~\cite{wen2016discriminative} and L-Softmax~\cite{liu2017sphereface} to reduce the complexity to $O(n)$. They lie in the assumption that data with the same label cluster around a certain center.

Face recognition is a typical application of metric learning, in which the similarity between certain two faces should be determined. Experiments on face recognition tasks show that CenterLoss and L-Softmax perform better than most other algorithms, which implies the effectiveness of unary losses over most pariwise/triplet ones in metric learning tasks. However, CenterLoss and L-Softmax lack theoretical guarantees that pairwise/triplet losses hold. Proxy-NCA~\cite{movshovitz2017no} proposes "proxies" to estimate a certain triplet loss and reduce the complexity of the metric learning, but the estimation is not accurate as it is greatly affected by the distances between the data and the "proxies".

\subsubsection{Semi-supervised Learning} Semi-supervised learning aims at learning with limited labeled data and huge amount of unlabeled data. Most semi-supervised learning methods lie in the smoothness assumption in which similar data are expected to share the same label~\cite{krause2010discriminative,weston2012deep}. These methods can be easily extended to semi-supervised hashing~\cite{yan2017semi,zhang2017ssdh}. But they are not working well for complex data as the smoothness assumption is not working well. Recently, Temporal Ensembling~\cite{laine2016temporal} proposes a perturbation based approach where a consensus prediction of a noisy input is formed. This method is further improved by adding more smoothness constraints like SNTG~\cite{luo2017smooth}, or performing ensembling on deep networks, denoting Mean Teacher~\cite{tarvainen2017mean}. These methods achieve great improvement on semi-supervised learning problems, and it is expected to utilize these methods to improve the semi-supervised hashing.

In this paper, we discover the Unary Upper Bound of the triplet losses, thus the classification-based unary loss and triplet losses are bridged. Moreover, we propose a novel hashing algorithm called Semantic Cluster Unary Loss(SCUL), which is based on the modified Unary Upper Bound. Then we extend it to the semi-supervised setting by combining the SCUL with the Mean Teacher. Experiments show its superiority over the state-of-the-art (semi-)supervised hashing algorithms.

\section{Unary Upper Bound for Supervised Hashing}
\label{sec:uub}

Suppose we are given $n$ data samples $\mathbf{x}_1, \mathbf{x}_2, ..., \mathbf{x}_n$, the goal of hash learning is to learn the hash function $H \revise{: \mathbf{x} \to \{-1,1\}^r}$, where $r$ is the code length. For supervised hashing, the semantic similarity information is crucial for learning the hashcodes, which is usually defined by whether the two data samples share certain semantic labels or tags.

\subsection{Revisiting Triplet Ranking Loss for Supervised Hashing}

Triplet ranking loss is widely used in the supervised (deep) hashing algorithms~\cite{lai2015simultaneous,norouzi2012hamming}. Given the training triplets $(\mathbf{x}, \mathbf{x}^+, \mathbf{x}^-)$ in which $\mathbf{x}, \mathbf{x}^+$ are semantically similar and $\mathbf{x}, \mathbf{x}^-$ are dissimilar, the most widely used triplet loss is
\begin{equation}
l_t(\mathbf{x}, \mathbf{x}^+, \mathbf{x}^-) = [m - |H(\mathbf{x}) - H(\mathbf{x}^-)| + |H(\mathbf{x}) - H(\mathbf{x}^+)|]_+
\end{equation}
where $[\cdot]_+ \doteq \max (0, \cdot)$, $|\cdot|$ is the distance measure (e.g. Hamming distance), and $m$ is the hyperparameter. It is expected to find a hash function where $H(\mathbf{x})$ is closer to $H(\mathbf{x}^+)$ than $H(\mathbf{x}^-)$. Another widely used triplet loss is NCA~\cite{goldberger2005neighbourhood}. \revise{It can be noticed that most triplet losses are monotonous, Lipschitz continuous functions~\cite{lai2015simultaneous,norouzi2012hamming,goldberger2005neighbourhood,dai2016binary}. In fact, the Lipschitz continuity is also widely applied in many machine learning problems like SVM, Logistic Regression, etc. As the gradient of the loss with Lipschitz continuity is constrained, the gradient descent procedure with these losses is expected to be stable and achieve good results.~\cite{arjovsky2017wasserstein}}

More generally, the triplet ranking loss can be formulated as
\begin{equation}
l_t(\mathbf{x}, \mathbf{x}^+, \mathbf{x}^-) = g(|H(\mathbf{x}) - H(\mathbf{x}^+)|, |H(\mathbf{x}) - H(\mathbf{x}^-)|)
\label{eq:trl}
\end{equation}
and $g(\cdot, \cdot)$ is a monotonous, Lipschitz continuous function such that
\begin{equation}
\begin{split}
g(a,b) &\ge 0 \\
0 \le g(a_2, b) - g(a_1, b) \le a_2 - a_1, \quad a_1 &\le a_2 \\
0 \le g(a, b_1) - g(a, b_2) \le b_2 - b_1, \quad b_1 &\le b_2 \\
\end{split}
\label{eq:g}
\end{equation}

Given $n$ training data samples $\mathbf{x}_1, \mathbf{x}_2 ..., \mathbf{x}_n$, denote $S$ as a set such that $(i,j) \in S$ implies $\mathbf{x}_i, \mathbf{x}_j$ are similar, the goal of hash learning is to optimize the following triplet loss function:
\begin{equation}
\min_{H} \mathcal{L}_t = \sum_{(i,j) \in S, (i,k) \notin S}  g(|\mathbf{h}_i - \mathbf{h}_j|, |\mathbf{h}_i - \mathbf{h}_k|)
\label{eq:trf}
\end{equation}
where $\mathbf{h}_i = H(\mathbf{x}_i), i = 1, 2, ..., n$ is the learned hashcode of $\mathbf{x}_i$.

\subsection{Unary Upper Bound for Triplet Ranking Loss}

As most supervised hashing problems, consider hashing on a dataset in which each data instance has a single semantic label. Denote $C$ as the number of semantic labels, and $y_1, ..., y_n \in \{1,2,...,C\}$ are the labels of $\mathbf{x}_1,..., \mathbf{x}_n$. The data pairs are similar if they share the same semantic label.

Intuitively, the intra-class variations among binary codes should be minimized while keeping inter-class distances far apart. \revise{We note that for algorithms where large amount of distance computations are involved like k-means, the complexities can be reduced by introducing some "centroids" for fast distance estimation.~\cite{elkan2003using} To what follows,} suppose there are $C$ auxiliary vectors $\mathbf{c}_1, ..., \mathbf{c}_C \in \mathbb{R}^r$, each of which corresponds to a certain semantic label. \revise{Considering the data triplet ($\mathbf{x}_i, \mathbf{x}_j, \mathbf{x}_k$) such that $y_i = y_j, y_i \neq y_k$, we have the following hamming distance estimation according to the triangle inequality:}
\begin{equation}
\begin{split}
|\mathbf{h}_i-\mathbf{h}_j| &\le |\mathbf{h}_i-\mathbf{c}_{y_i}| + |\mathbf{h}_j-\mathbf{c}_{y_j}|, \quad y_i=y_j \\
|\mathbf{h}_i-\mathbf{h}_k| &\ge |\mathbf{h}_i-\mathbf{c}_{y_k}| - |\mathbf{h}_k-\mathbf{c}_{y_k}|, \quad y_i \neq y_k
\end{split}
\label{eq:tri}
\end{equation}

\begin{figure}[t]
    \setlength{\abovecaptionskip}{2pt}
    \setlength{\belowcaptionskip}{0pt}
    \centering
    \includegraphics[scale=0.4]{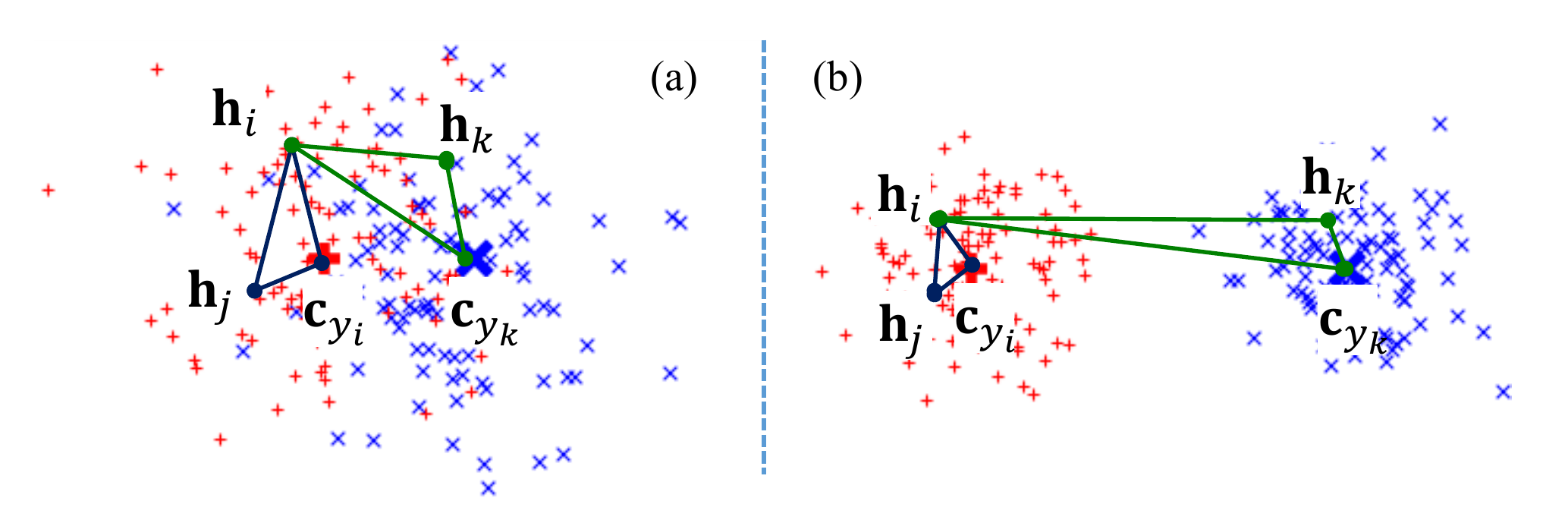}
    \caption{Illustration on \revise{the relationships of the codes and the cluster centers as well as on} training the Unary Upper Bound. $\mathbf{h}_i, \mathbf{h}_j, \mathbf{h}_k$ and their relationships are defined in Eq. (\ref{eq:tri}). \revise{The upper bound of $|\mathbf{h}_i-\mathbf{h}_j|$ can be arrived at with auxiliary vector $\mathbf{c}_{y_i}$ and the lower bound of $|\mathbf{h}_i-\mathbf{h}_k|$ can be obtained by $\mathbf{c}_{y_k}$.} (a) Before training: the codes of two classes are not separated apart, so the triplet loss and the Unary Upper Bound is large. (b) After training: two cluster centers are far apart, the intra-class distances are small, thus the triplet loss and the Unary Upper Bound are small.}
    \label{fig:uub}
\end{figure}

\revise{Eq. (\ref{eq:tri}) arrives at an upper bound of intra-class distances of hashcodes and a lower bound of the inter-class distances, and the illustration is shown in Figure \ref{fig:uub}. It should be noticed that $|\mathbf{h}_i-\mathbf{h}_k| \ge \big| |\mathbf{h}_i-\mathbf{c}_{y_k}| - |\mathbf{h}_k-\mathbf{c}_{y_k}| \big|$ holds for any $\mathbf{h}_i, \mathbf{h}_k, \mathbf{c}_{y_k}$ according to the inverse triangle inequality, and the second inequality of Eq. (\ref{eq:tri}) holds regardless of the absolute value. As discussed below, removing the absolute value has no influence on the correctness of the lower bound of inter-class distances.} 

With the property of $g(\cdot,\cdot)$ shown in Eq. (\ref{eq:g}), we can arrive at an upper bound of triplet ranking loss: 
\begin{equation}
\setlength{\abovedisplayskip}{3pt}
\setlength{\belowdisplayskip}{3pt}
\begin{split}
g(&|\mathbf{h}_i - \mathbf{h}_j|, |\mathbf{h}_i - \mathbf{h}_k|) \\
&\le g(|\mathbf{h}_i-\mathbf{c}_{y_i}| + |\mathbf{h}_j-\mathbf{c}_{y_j}|, |\mathbf{h}_i-\mathbf{c}_{y_k}| - |\mathbf{h}_k-\mathbf{c}_{y_k}|) \\
&\le g(|\mathbf{h}_i-\mathbf{c}_{y_i}|, |\mathbf{h}_i-\mathbf{c}_{y_k}|) + (|\mathbf{h}_j-\mathbf{c}_{y_j}|+|\mathbf{h}_k-\mathbf{c}_{y_k}|) \\
& (y_i = y_j, y_i \neq y_k)
\end{split}
\label{eq:triest}
\end{equation}
thus the triplet ranking loss can be represented by the distances between the hashcodes and the $C$ auxiliary vectors $\mathbf{c}_1, ..., \mathbf{c}_C$.

If the class labels are evenly distributed (for datasets with unbalanced labels, the labels can be balanced by sampling), the upper bound of Eq. (\ref{eq:trf}) can be arrived by a simple combination of triplet losses shown in Eq. (\ref{eq:triest}) such that
\begin{equation}
\mathcal{L}_t \le (\frac{n}{C})^2 (C-1) \sum_{i=1}^n[ l_c(\mathbf{h}_i, y_i) +  2 |\mathbf{h}_i-\mathbf{c}_{y_i}| ]
\label{eq:uub}
\end{equation}
where $l_c(\mathbf{h}_i, y_i) = \frac{1}{C-1}\sum_{l=1,l \neq y_i}^C g(|\mathbf{h}_i-\mathbf{c}_{y_i}|, |\mathbf{h}_i-\mathbf{c}_{l}|)$ can be regarded as max-margin multiclass classification loss such as multi-class hinge loss
\begin{equation}
l_c(\mathbf{h}_i, y_i) = \frac{1}{C-1}\sum_{l=1,l \neq y_i}^C [m + |\mathbf{h}_i-\mathbf{c}_{y_i}| - |\mathbf{h}_i-\mathbf{c}_{l}|]_+
\end{equation}
or softmax loss
\begin{equation}
\begin{split}
&l_c(\mathbf{h}_i, y_i) = - \log \frac{\exp (-|\mathbf{h}_i-\mathbf{c}_{y_i}|)}{\sum_{j=1}^C \exp (-|\mathbf{h}_i-\mathbf{c}_{j}|)} \\
=& \frac{1}{C-1} \sum_{l=1,l \neq y_i}^C [ -\log \frac{\exp (-|\mathbf{h}_i-\mathbf{c}_{y_i}|)}{\substack{e^ {-|\mathbf{h}_i-\mathbf{c}_{y_i}|} + e^ {-|\mathbf{h}_i-\mathbf{c}_l|} + \sum_{\substack{j=1 \\ j\neq y_i \\ j\neq l}}^C e^{-|\mathbf{h}_i-\mathbf{c}_{j}|}}} ]
\end{split}
\end{equation}

The derivation of Eq (\ref{eq:uub}) is shown in the supplemental material. We name the right side of Eq. (\ref{eq:uub}) as the Unary Upper Bound of the triplet loss.

\subsection{Unary Upper Bound and Semantic Cluster Unary Loss}
\label{sec:duub}

From Eq. (\ref{eq:uub}) we can see that the complexity of the Unary Upper Bound is just $O(n)$. It is clear that $\mathbf{c}_1, ..., \mathbf{c}_C$ in Eq. (\ref{eq:uub}) can be regarded as $C$ cluster centers, and each cluster corresponds to a certain semantic label. If the distance between the data instance and the corresponding cluster center is small, the intra-class distances are expected to be small; and if the data instance and other centers are separated far apart and the intra-class distances are small, the inter-class distances are expected to be large. Thus we can arrive at an alternative way to minimize the triplet loss: (1) minimize the distance between the data instance $\mathbf{h}_i$ and the corresponding cluster center $\mathbf{c}_{y_i}$ by minimizing $l_c(\mathbf{h}_i, y_i)$ and $|\mathbf{h}_i-\mathbf{c}_{y_i}|$; (2) maximize the distance between $\mathbf{h}_i$ and other cluster centers by minimizing $l_c(\mathbf{h}_i, y_i)$. The training procedure is shown in Figure \ref{fig:uub}. After optimization, $C$ clusters can be formed and each cluster corresponds to a certain semantic label, and we name each of them as the {\em Semantic Cluster}. The {\em Semantic Cluster} centers can be regarded as $\mathbf{c}_1, ..., \mathbf{c}_C$.

However, the Unary Upper Bound in Eq. (\ref{eq:uub}) is too loose, and the direct minimization of the Unary Upper Bound may suffer from collapse. As the $|\mathbf{h}_i-\mathbf{c}_{y_i}|$ term takes an crucial part in the Unary Upper Bound, it is possible that $\mathbf{c}_{y_i}, \mathbf{h}_i$ will converge to zero to ensure that $|\mathbf{h}_i-\mathbf{c}_{y_i}|=0$, then the Unary Upper Bound is relatively small, but the original triplet loss is large. The loose of the Unary Upper Bound is caused in two aspects: (1) In the beginning, the bounds in Eq. (\ref{eq:tri}) is too loose \revise{and the lower bound in the second inequality may be smaller than zero}, thus the first inequality of Eq. (\ref{eq:triest}) is loose; (2) at the end of optimization, the triplet loss will goes to zero, thus $g(\cdot,\cdot) \to 0, g'_a(a, \cdot) \to 0, g'_b(\cdot, b) \to 0$, and it is clear that the second inequality of Eq. (\ref{eq:triest}) will be loose.

\begin{figure}[t]
    \setlength{\abovecaptionskip}{2pt}
    \setlength{\belowcaptionskip}{0pt}
    \centering
    \includegraphics[scale=0.35]{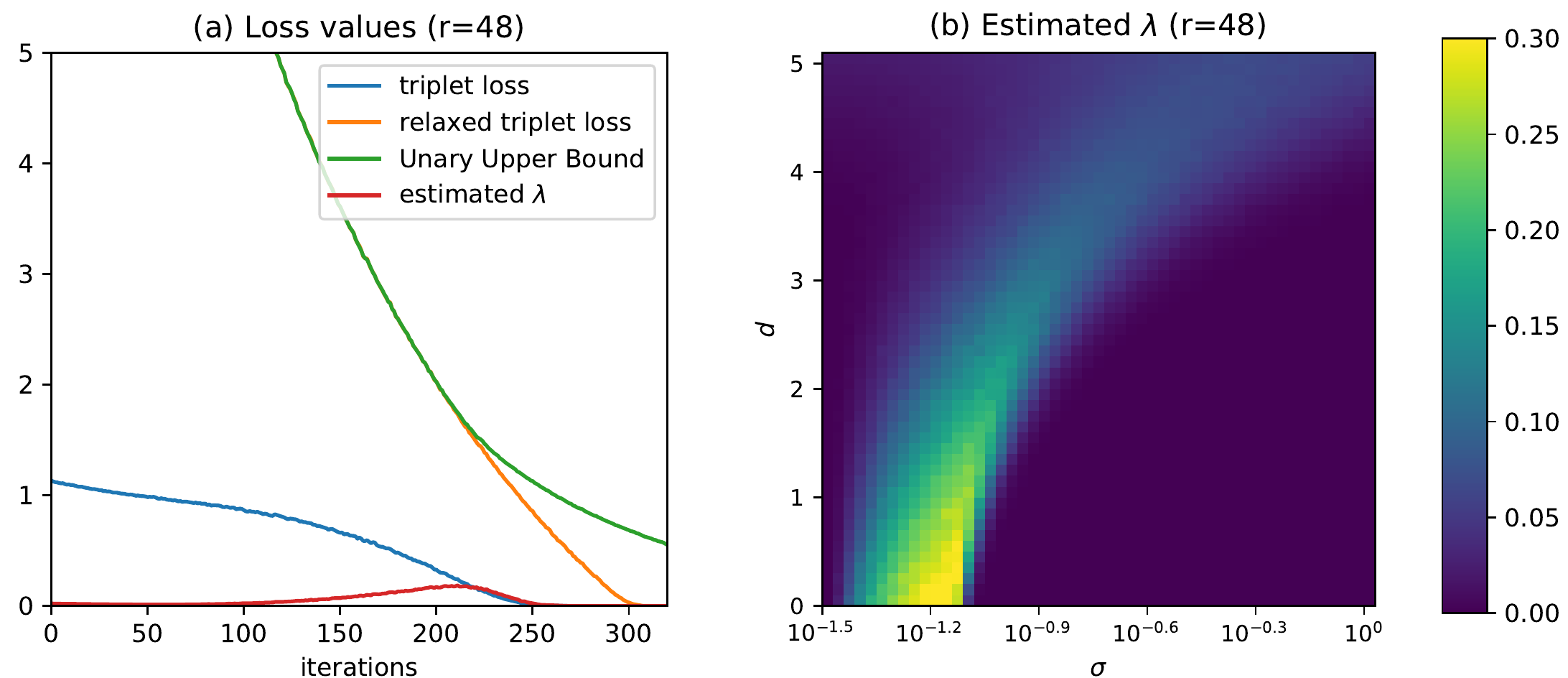}
    \caption{Illustration on the relationship between the triplet loss and the Unary Upper Bound, as well as the estimated $\lambda$ defined in Eq. (\ref{eq:tuub}). Denote $\sigma$ as the variance of the cluster and $d$ as the distance between clusters. (a) The change of different values as $\sigma$ decreases and $d$ increases. (b) Value of $\lambda$ with varied $\sigma$ and $d$.}
    \label{fig:illus-uub}
\end{figure}

To address this issue, we want to find a tighter bound such that
\begin{equation}
\mathcal{L}_t \le M_t \mathcal{L}_u \quad \mathcal{L}_u = \sum_{i=1}^n [ l_c(\mathbf{h}_i, y_i) +  \lambda |\mathbf{h}_i-\mathbf{c}_{y_i}| ]
\label{eq:tuub}
\end{equation}
where $M_t = (\frac{n}{C})^2 (C-1)$ and $\lambda > 0$. The analytic form of $\lambda$ is intractable due to the unknown type of $g$ and the distribution of $\mathbf{h}_i$. We propose a toy example to show that $\lambda$ can be quite small. Denote $\mathbf{h}_i \in \mathbb{R}^r, r=48$ in this example. Suppose there are $C=2$ clusters. Each cluster corresponds to a Gaussian distribution with covariance $\sigma \mathbf{I}_r$, and the distance between cluster centers (the means of Gaussian distributions) is $d$. The triplet ranking loss is defined as $g(a, b) = [a-b+1]_+$. During the training process, it is expected that $\sigma$ gradually decreases and $d$ gradually increases. Figure \ref{fig:illus-uub}(a) shows the change of the original triplet loss, the Unary Upper Bound, and the triplet loss defined by the second term of Eq. (\ref{eq:triest}) (denote relaxed triplet loss). An estimation of $\lambda \doteq (\mathcal{L}_t / M_t - \sum_{i=1}^n l_c(\mathbf{h}_i, y_i) )/ \sum_{i=1}^n |\mathbf{h_i} - y_i|$ is also computed. It can be seen clearly that all losses are gradually decreasing and $\lambda$ is small during the whole process. Figure \ref{fig:illus-uub}(b) illustrates the estimated $\lambda$ with varied $\sigma$ and $d$, which implies that $\lambda$ is relatively small under almost all conditions. Thus it is expected that a tighter Unary Upper Bound as Eq. (\ref{eq:tuub}) exists with a relatively small $\lambda$.

To conclude, the modified form of the Unary Upper Bound of the triplet loss can be written as Eq. (\ref{eq:tuub}), and we name it as {\em Semantic Cluster Unary Loss(SCUL)}. By optimizing the SCUL in Eq. (\ref{eq:tuub}), not only a smaller $|\mathbf{h}_i-\mathbf{c}_{y_i}|$ but also a larger $|\mathbf{h}_i-\mathbf{c}_{y_k}|, y_i \neq y_k$ can be achieved, so that the original triplet loss is able to be minimized. Moreover, optimizing Eq. (\ref{eq:tuub}) is expected to be efficient as the complexity is just $O(n)$.

\begin{figure*}[t]
    \setlength{\abovecaptionskip}{0pt}
    \setlength{\belowcaptionskip}{0pt}
    \centering
    \includegraphics[scale=0.45]{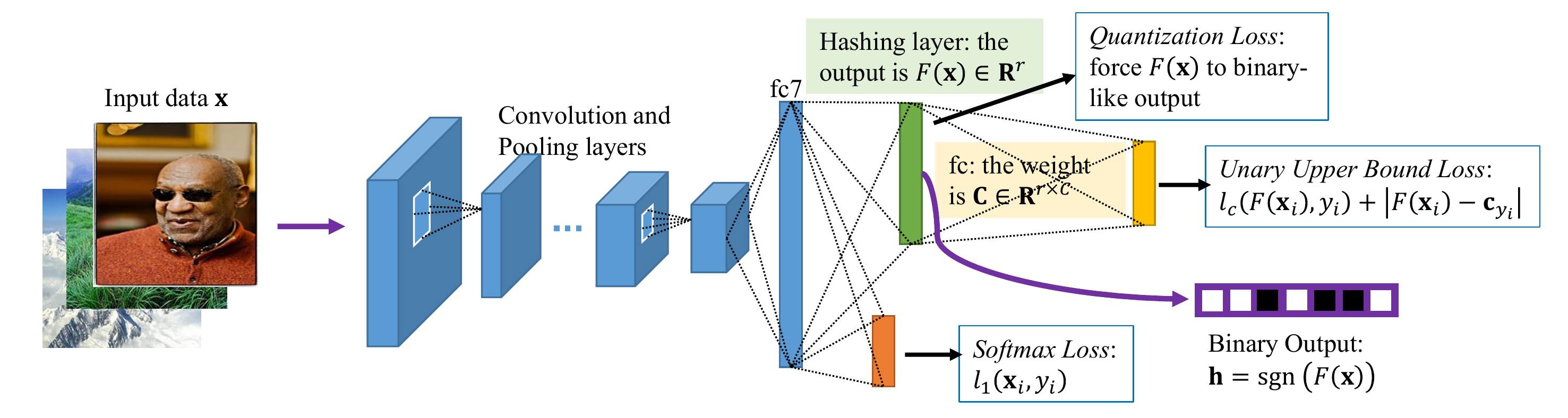}
    \caption{Overview of the {\em Semantic Cluster Deep Hashing} algorithm. Given input data $\mathbf{x}$, the hash value can be obtained from the hashing layer (green rectangle) with $\mathbf{h} = \mathrm{sgn} (F(\mathbf{x}))$, where $F(\mathbf{x})$ is the activation of the hashing layer. $\mathbf{C} \in \mathbb{R}^{r \times C}$ is the parameter of the last fully-connected layer before the SCUL, and the columns of $\mathbf{C}$ can be regarded as $C$ cluster centers. This network is jointly trained by the SCUL, the quantization loss and the softmax loss, defined in Eq. (\ref{eq:robj}).}
    \label{fig:framework}
\end{figure*}

\subsection{Multilabel Extension}

In the previous section, we have just considered hashing on data with single semantic label. In practical applications, lots of data have more than one semantic label, and the similarity is defined by whether two data instances share certain amount of labels. Hashing on data with multilabel should be considered.

Consider a case where semantic labels are evenly distributed. In particular, the probability of each $\mathbf{x}_i, i=1,...,n$ has a certain label $l \in \{1,...,C\}$ is $p$. Denote that the similarity is defined by whether two data instances share at least one semantic label, and the triplet ranking loss is defined by
\begin{equation}
\mathcal{L}_{mt} = \sum_{(i,j) \in S, (i,k) \notin S} r_{ij} g(|\mathbf{h}_i - \mathbf{h}_j|, |\mathbf{h}_i - \mathbf{h}_k|)
\label{eq:mtrs}
\end{equation}
where $r_{ij} \ge 1$ denotes the number of labels $\mathbf{x}_i$ and $\mathbf{x}_j$ share. Note that we attach greater importance to similar data pairs with more shared semantic labels. Then we can arrive at an Unary Upper Bound of the above triplet ranking loss, which is concluded in the following proposition:
\begin{prop}
Denote $Y_i \subseteq \{1,2,...,C\}$ as the labels of data instance $\mathbf{x}_i$, and $\mathrm{P}(l \in Y_i) = p$ for all $l = 1,2,...,C$. The Unary Upper Bound of the expectation value of the triplet loss defined by Eq. (\ref{eq:mtrs}) is
\begin{equation}
\begin{split}
\mathbb{E}[\mathcal{L}_{mt}] \le & (C-1)p^2 n^2 \sum_{i=1}^n [q(|Y_i|) l_{mc}(\mathbf{h}_i, Y_i) \\
&+ (Q+q(|Y_i|)) \sum_{s \in Y_i} |\mathbf{h}_i - \mathbf{c}_s|]
\end{split}
\end{equation}
where $q(x) = \frac{C-x}{C-1}(1-p)^x, Q = (1-p)^2(1-p^2)^{C-2}$, $|Y_i|$ denotes the number of labels $\mathbf{x}_i$ contains, and
\begin{equation}
l_{mc}(\mathbf{h}_i, Y_i) = \frac{1}{C-|Y_i|} \sum_{s \in Y_i} \sum_{t \notin Y_i} g(|\mathbf{h}_i - \mathbf{c}_s|, |\mathbf{h}_i - \mathbf{c}_t|)
\end{equation}
can be regarded as a multilabel softmax loss such that
\begin{equation}
l_{mc}(\mathbf{h}_i, Y_i) = \sum_{s \in Y_i} [ -\log \frac{\exp (-|\mathbf{h}_i-\mathbf{c}_{s}|)}{\sum_{j=1}^C \exp(-|\mathbf{h}_i-\mathbf{c}_{j}|)} ]
\label{eq:msl}
\end{equation}
\label{prop:muub}
\end{prop}

The proof of this proposition is shown in the supplemental material. Similar as Section \ref{sec:duub}, we can arrive at a more general form such that
\begin{equation}
\begin{split}
\mathbb{E}[\mathcal{L}_{mt}] \le M_{mt} \mathcal{L}_{mu} \\
\mathcal{L}_{mu} = \sum_{i=1}^n [q(|Y_i|) l_{mc}(\mathbf{h}_i, Y_i) + u(|Y_i|) \sum_{s \in Y_i} |\mathbf{h}_i - \mathbf{c}_s|]
\end{split}
\label{eq:mtuub}
\end{equation}
where $M_{mt}$ is a constant and the value of $u(x)$ is relatively small. Eq. (\ref{eq:mtuub}) can be regraded as the multilabel version of SCUL and the complexity is also reduced to $O(n)$. 

\section{Semantic Cluster Deep Hashing}
\label{sec:scdh}

In this section, we propose a novel deep supervised hashing algorithm called Semantic Cluster Deep Hashing (SCDH), in which the Semantic Cluster Unary Loss(SCUL) in Eq. (\ref{eq:tuub}) is adopted as the loss to optimize. The term "Semantic Cluster" is defined in Sec \ref{sec:duub}. The proposed algorithm is expected to be efficient, as the classification-based SCUL is introduced.

\subsection{Overall Architecture}

The overall network architecture is shown in Figure \ref{fig:framework}. Denote $\mathrm{fc7}$ as the last but one layer of a classification network (eg. AlexNet, VGGNet, etc.), there are two ways after $\mathrm{fc7}$. One way consists of two fully connected layers without non-linear activations. The first layer is the hashing layer with $r$ outputs, and the second layer has $C$ outputs. $\mathbf{C} = [\mathbf{c}_1, ..., \mathbf{c}_C] \in \mathbb{R}^{r \times C}$ is the parameters of the second layer, which can be regarded as $C$ cluster centers. The other way is a fully-connected layer(denote $\mathrm{fc8}$) with the softmax classification loss. 

Denote $F(\mathbf{x})$ as the activations of the hashing layer, and $H(\mathbf{x}) = \mathrm{sgn} (F(\mathbf{x}))$ as the hash function, where $\mathrm{sgn}$ is the element-wise sign function and $F(\mathbf{x}) \in \mathbb{R}^r$. The objective is learning $F$ and $\mathbf{C}$ by optimizing the SCUL to obtain a good hash function. 

After training the network, the binary codes of data $\mathbf{x}$ are easily obtained with $\mathbf{h} = \mathrm{sgn} (F(\mathbf{x}))$.

\subsection{Loss Function}

Denote $\mathbf{h}_i = \mathrm{sgn} (F(\mathbf{x}_i)), i=1,2,...,n$, we should optimize the SCUL such that
\begin{equation}
\min_{F, \mathbf{C}} \mathcal{L}_u = \sum_{i=1}^n [ l_c(\mathbf{h}_i, y_i) + \lambda |\mathbf{h}_i - \mathbf{c}_{y_i}| ]
\label{eq:inituub}
\end{equation} 

We regard $|\cdot|$ as the euclidean distance in Eq. (\ref{eq:inituub}). As discussed before, the complexity of the proposed loss is just $O(n)$ and the loss has theoretical relationship with the triplet ranking loss. Similar as~\cite{guo2016hash,xia2014supervised}, For faster convergence, we add another classification loss $\mathcal{L}_1 = \sum_{i=1}^n l_1(\mathbf{x}_i, y_i)$ to train the deep neural network:
\begin{equation}
\min_{F, \mathbf{C}} \mathcal{L} = \mathcal{L}_u + \mu \mathcal{L}_1 = \sum_{i=1}^n [ l_c(\mathbf{h}_i, y_i) + \mu l_1(\mathbf{x}_i, y_i) + \lambda |\mathbf{h}_i - \mathbf{c}_{y_i}| ]
\label{eq:obj}
\end{equation}

Inspired by deep methods for classification problems, for hashing on data with single label, $l_c(\mathbf{h}_i, y_i)$ has a similar form with softmax loss such that $l_c(\mathbf{h}_i, y_i) = -\log \frac{\exp (-|\mathbf{h}_i-\mathbf{c}_{y_i}|)}{\sum_{j=1}^C \exp(-|\mathbf{h}_i-\mathbf{c}_{j}|)} $. 

For datasets with multilabels, $l_c(\mathbf{h}_i, y_i) + \lambda |\mathbf{h}_i - \mathbf{c}_{y_i}|$ can be replaced by $q(|Y_i|) l_{mc}(\mathbf{h}_i, Y_i) + \lambda \sum_{s \in Y_i} |\mathbf{h}_i - \mathbf{c}_s|$, where $l_{mc}(\mathbf{h}_i, Y_i)$ is defined in Eq. (\ref{eq:msl}) and we simply use $q(x) = 1/x$. $l_1(\mathbf{x}_i, y_i)$ is the multilabel softmax loss.

Note that $\mathcal{L}_u$ has a similar formulation with CenterLoss~\cite{wen2016discriminative} to some extent, but the differences between two losses are obvious. First, CenterLoss just use softmax loss after the feature embedding layer, and the centers in the CenterLoss have no relationship with the last fully-connected layer. Second, our proposed $\mathcal{L}_u$ has the theoretical relationship with the triplet ranking loss. Furthermore, SCDH has a simple but novel extension of the multilabel case.

\subsection{Relaxation}
\label{sec:relax}

\begin{figure}[t]
    \centering
    \includegraphics[scale=0.5]{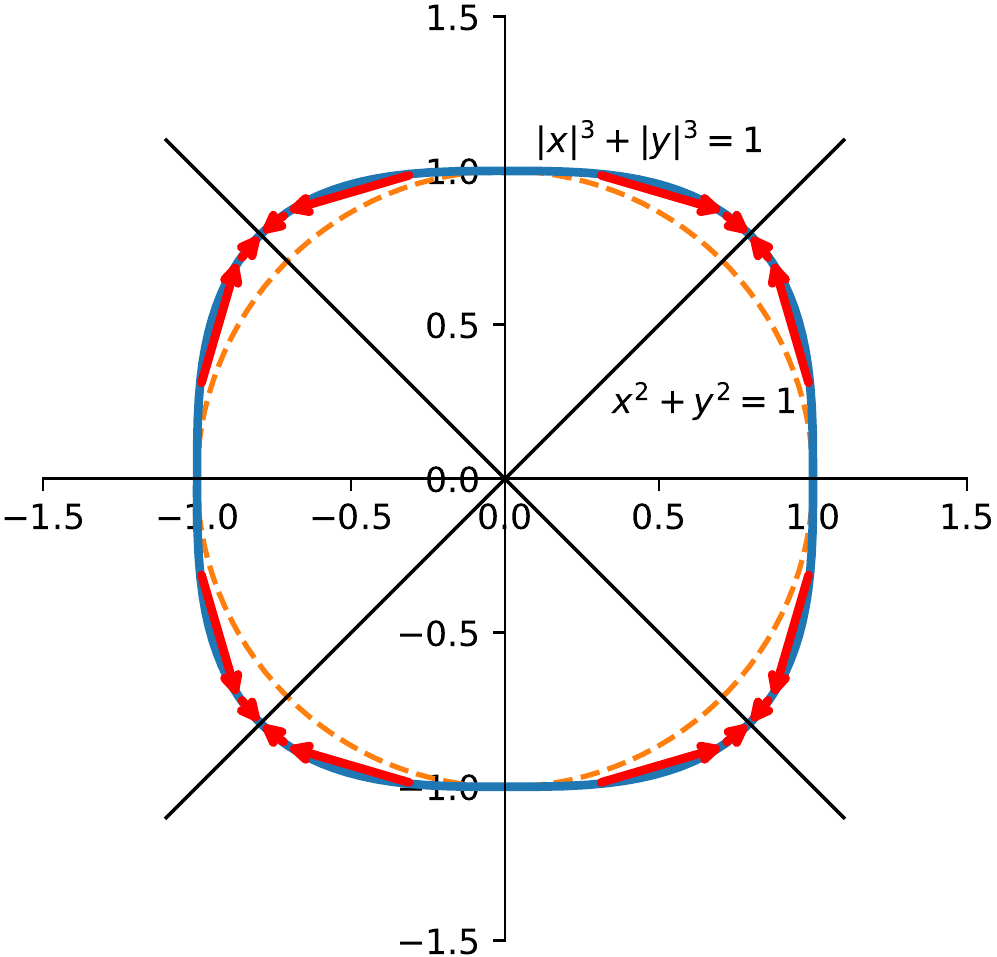}
    \caption{\revise{Illustration on the effect of optimizing with the proposed quantization loss defined in Eq. (\ref{eq:robj}). The blue circle denotes a set of 3-normed points. The red arrows denote the gradients of the proposed quantization loss. It can be seen clearly that each element of the vector $F(\mathbf{x})$ will be almost the same during training, so $F(\mathbf{x})$ and $\mathrm{sgn}(F(\mathbf{x}))$ are expected to be close to each other. A unit circle (2-norm) is shown in this figure for comparison.}}
    \label{fig:ql}
\end{figure}

The main difficulty in optimizing Eq. (\ref{eq:obj}) is the existence of the discrete constraints, making it intractable to train the network with back-propagation. Recent researches convey that removing the discrete constraints as well as adding the quantization loss is a good approach~\cite{zhu2016deep,li2015feature,cao2016deep} in which the activation of the hashing layer $F(\mathbf{x})$ is not only continuous but also around $+1/-1$. \revise{However, optimizing these losses has to make the norm of $F(\mathbf{x})$ constrained. In fact, we just need to push the elements of the learned $F(\mathbf{x})$ away from zero so that less discrepancy is involved after generating the codes with the $\mathrm{sgn}$ function. For a well-learned $F(\mathbf{x})$ with relatively small or large norm, we may not necessarily optimize with the traditional quantization loss.~\cite{cao2016correlation}}


In this paper, we introduce a new quantization loss such that $l_q(\mathbf{f}) =  1- \frac{\mathbf{1}^\mathrm{T} \mathrm{abs}(\mathbf{f})}{ \| \mathbf{1} \|_q \| \mathbf{f} \|_p }$ and arrive at the following relaxed problem:
\begin{equation}
\begin{split}
\min_{F, \mathbf{C}} \mathcal{L} =& \sum_{i=1}^n [ l_c(F(\mathbf{x}_i), y_i) + \mu l_1(\mathbf{x}_i, y_i)  \\ 
+& \lambda |F(\mathbf{x}_i) - \mathbf{c}_{y_i}| + \alpha l_q(F(\mathbf{x}_i)) ]
\end{split}
\label{eq:robj}
\end{equation}
where $\mathrm{abs}(\cdot)$ is the element-wise absolute function, $\| \cdot \|_p, \| \cdot \|_q$ form a pair of dual norms such that $1/p+1/q=1$, and therefore $\mathrm{abs}(\mathbf{x})^\mathrm{T} \mathrm{abs}(\mathbf{y}) \le \| \mathbf{x} \|_p \| \mathbf{y} \|_q $ for any $\mathbf{x}, \mathbf{y}$ \revise{according to the Holder's inequality. As shown in Figure \ref{fig:ql}}, by optimizing the proposed quantization loss, each element of $F(\mathbf{x})$ is expected to be almost the same, thus less discrepancy between $F(\mathbf{x})$ and $\mathrm{sgn} (F(\mathbf{x}))$ will be involved. Moreover, unlike the quantization losses proposed in~\cite{zhu2016deep,cao2016deep}, ours do not need to constrain the norm of $F(\mathbf{x})$ during optimization.

It is easier to optimize the quantization loss with greater $p$. We use $p=3, q=1.5$ in this algorithm.

\subsection{Optimization}
\label{sec:optimization}

It it clear that the proposed SCDH can be trained end-to-end with back-propagation, in which Eq. (\ref{eq:robj}) can be optimized with gradient descent. The crucial term in Eq. (\ref{eq:robj}) for gradient descent is the SCUL term, denoting $l_u(\mathbf{x}_i) = l_c(F(\mathbf{x}_i), y_i) + \lambda |F(\mathbf{x}_i) - \mathbf{c}_{y_i}|$, and the gradient is computed as
\begin{equation}
\begin{split}
\nabla_{\mathbf{c}_{y_i}} l_u =& (1-p_{y_i} + \lambda) \nabla_{\mathbf{c}_{y_i}} |F(\mathbf{x}_i) - \mathbf{c}_{y_i}| \\
\nabla_{\mathbf{c}_{j}} l_u =& -p_j \nabla_{\mathbf{c}_{j}} |F(\mathbf{x}_i) - \mathbf{c}_{j}| \quad j \neq y_i \\
\nabla_{F(\mathbf{x}_i)} l_u =& (1+\lambda) \nabla_{F(\mathbf{x}_i)} |F(\mathbf{x}_i) - \mathbf{c}_{y_i}| \\
&- \sum_{j=1}^C p_j \nabla_{F(\mathbf{x}_i)} |F(\mathbf{x}_i) - \mathbf{c}_{j}|
\end{split}
\label{eq:gruub}
\end{equation}
where $p_k = \frac{\exp (-|\mathbf{h}_i-\mathbf{c}_{k}|)}{\sum_{j=1}^C \exp(-|\mathbf{h}_i-\mathbf{c}_{j}|)}$ can be regarded as the probability, and $\nabla |F(\mathbf{x}_i) - \mathbf{c}_{j}| = (F(\mathbf{x}_i) - \mathbf{c}_{j}) / |F(\mathbf{x}_i) - \mathbf{c}_{j}| $ is the gradient of the distance. By using Eq. (\ref{eq:gruub}), the network is able to be optimized end-to-end with back-propagation.

However, when training the dataset with a large number of classes like Imagenet, we find that the optimization of Eq. (\ref{eq:robj}) may be hard to converge. We find that $p_k, k=1,...,C$ will be relatively small in the beginning, especially on large number of classes, thus $\nabla_{\mathbf{c}_{y_i}} |F(\mathbf{x}_i) - \mathbf{c}_{y_i}|, \nabla_{F(\mathbf{x}_i)} |F(\mathbf{x}_i) - \mathbf{c}_{y_i}|$ will take the crucial part in computing the gradients. In this case, $F(\mathbf{x}_i), \mathbf{c}_{y_i}$ may be encouraged to go to zero to make $|F(\mathbf{x}_i) - \mathbf{c}_{y_i}| = 0$, and we pay less attention to enlarging the inter-class distances, making the optimization hard to converge or even collapse.

To overcome the above issue, we propose a {\em warm-up} procedure before training directly with back-propagation. In the beginning epochs, we simply constrain the Semantic Cluster centers $\mathbf{C}$ to a certain norm $s$ to avoid the centers and $F(\mathbf{x}_i)$ to go zero. After the {\em warm-up} procedure, $p_{y_i}$ goes larger, then the importance of $\nabla|F(\mathbf{x}_i) - \mathbf{c}_{y_i}|$ is diminished and $F(\mathbf{x}_i), \mathbf{c}_{y_i}$ is hard to go zero. Moreover, a larger $F(\mathbf{x}_i)$ makes it easier to enlarge the inter-class distances. For relatively small datasets, we do not need to use the {\em warm-up} procedure, but for faster convergence, the centers should be randomly initialized such that the norms of the centers should be large enough to prevent them going zero.

\begin{figure}[t]
    \setlength{\abovecaptionskip}{1pt}
    \setlength{\belowcaptionskip}{0pt}
    \centering
    \includegraphics[scale=0.65]{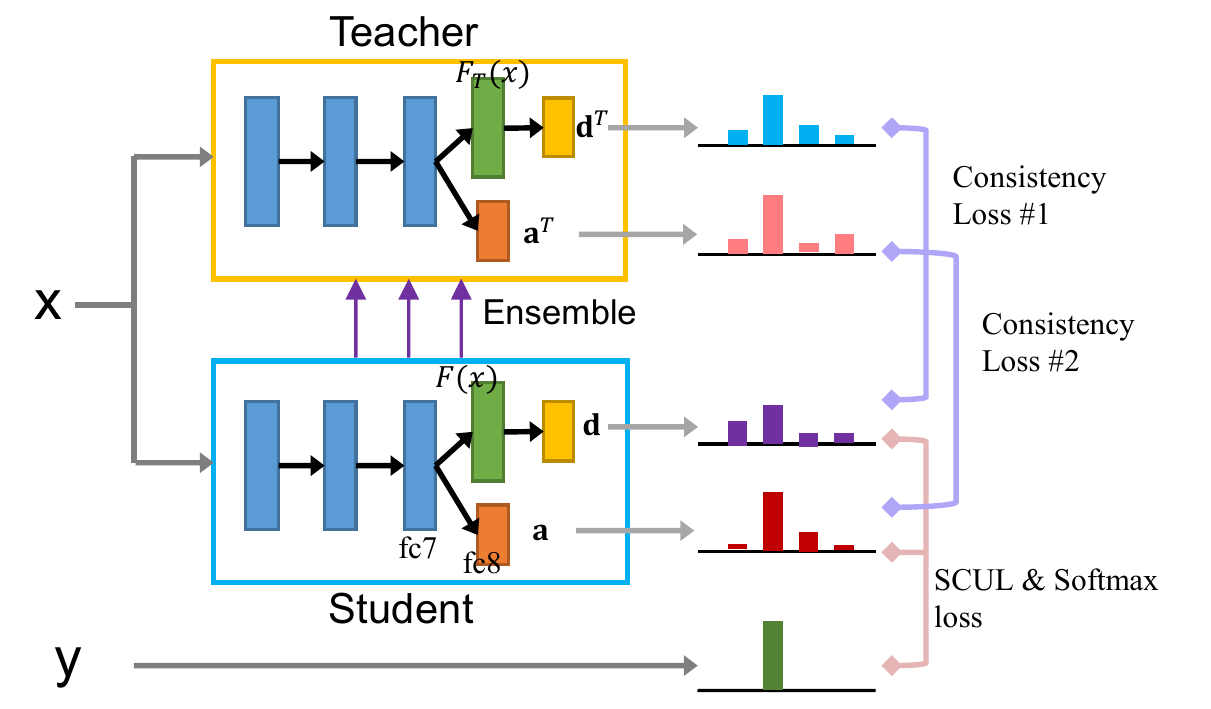}
    \caption{Overview of the MT-MCDH algorithm. The structure of teacher/student network are the same as SCDH. Different colors denote different types of layers, which are the same with that in Figure \ref{fig:framework}.}
    \label{fig:mtscdh-framework}
\end{figure}

\section{Mean Teacher Based Semi-Supervised Hashing}
\label{sec:ssh}

In this section, we extend the proposed SCDH algorithm for semi-supervised hashing by combining the proposed SCUL with Mean Teacher(MT)~\cite{tarvainen2017mean}. We name the proposed algorithm as {\em MT-SCDH}.

\subsection{Mean Teacher Recap}

Mean Teacher~\cite{tarvainen2017mean} is the current state-of-the-art algorithm for semi-supervised learning. It assumes that the data predictions should be consistent under different perturbations, so that it can satisfy the smoothness assumption. In this case, Mean Teacher proposes a dual role, i.e., the teacher and the student. The student is learned as before; the teacher is the average of consecutive student models in which the weights are updated as an exponential moving average(EMA) of the student weights. In addition, the outputs of the teacher are regarded as targets for the training the student, so we can introduce the following consistency loss as regularization for training:
\begin{equation}
\mathcal{R}_c = \sum_{i \in \mathcal{S} \cup \mathcal{U}} d(f(\tilde{\mathbf{x}}_i^{(1)}), f_T(\tilde{\mathbf{x}}_i^{(2)}))
\end{equation}
where $\mathcal{S}, \mathcal{U}$ are the labeled and unlabeled datasets respectively, $\tilde{\mathbf{x}}^{(1)},\tilde{\mathbf{x}}^{(2)}$ are two random perturbations of the original data point $\mathbf{x}$, $f(\cdot), f_T(\cdot)$ are the outputs of student or teacher network respectively, and $d(\cdot,\cdot)$ is the distance between two features. In particular, the teacher can be regarded as the average ensemble of the student, thus $f_T(\cdot)$ can be regarded as the mean of the student's outputs, making the training of the consistency loss more stable.

\subsection{The MT-SCDH Model}

As the proposed SCUL is a classification-based unary loss, it can be easily incorporated with the Mean Teacher for semi-supervised hashing in MT-SCDH. MT-SCDH is also a teacher-student model which is shown in Figure \ref{fig:mtscdh-framework}. The student is learned in the same manner as SCDH with the labeled dataset. The teacher is updated in the same way as the Mean Teacher. For all training data, we apply the consistency loss among the outputs of the networks.

Denote $\mathcal{S} = \{(\mathbf{x}_1, y_1), ..., (\mathbf{x}_n, y_n) \}$ as the labeled data and $\mathcal{U} = \mathbf{x}_{n+1}, ..., \mathbf{x}_m$ as the unlabeled data. The learning problem of MT-SCDH is shown as follows:
\begin{equation}
\begin{split}
\min_{F, \mathbf{C}} \mathcal{L}^s = \mathcal{L} &+ w \sum_{i \in \mathcal{S} \cup \mathcal{U}} [ \mu d(\mathbf{a}_i, \mathbf{a}_i^T) + d(\mathbf{d}_i, \mathbf{d}_i^T) ]
\\ &+ \alpha \sum_{j \in \mathcal{U}} l_q(F(\mathbf{x}_j))
\end{split}
\end{equation}
where $d(\cdot, \cdot) = \| \mathrm{softmax}(\cdot) - \mathrm{softmax}(\cdot) \|^2$, $\mathcal{L}$ is the supervised term defined in Eq. (\ref{eq:robj}), $l_q(\cdot)$ is the quantization loss proposed in Sec. \ref{sec:relax}, $w$ is the weight of the consistency loss, $\mathbf{a}_i, \mathbf{a}_i^T$ are the outputs of $\mathrm{fc8}$ for $\mathbf{x}_i$ in the  student and teacher network respectively, $\mathbf{d}_i, \mathbf{d}_i^T$ denote the negative distances between $F(\mathbf{x}_i)$ and the semantic clusters in the student and teacher network respectively. In other words, $\mathbf{d}_i = -[|F(\mathbf{x}_i)-\mathbf{c}_1|,...,|F(\mathbf{x}_i)-\mathbf{c}_C|]^\mathrm{T}$, and so as $\mathbf{d}_i^T$. Note that the input $\mathbf{x}_i$ is the random perturbations.

The basic configurations and training procedure are almost the same as SCDH. And we follow the suggested configurations (e.g. the value of $w$) in~\cite{tarvainen2017mean} for semi-supervised learning.

After training the network. we can extract the codes from either the teacher network or the student network.

\section{Experiments}
\label{sec:experiments}

In this section, we conduct various large-scale retrieval experiments to show the efficiency of the proposed SCDH methods. We compare our SCDH method with recent state-of-the-art (semi)-supervised deep hashing methods on the retrieval performance and the training time. Some ablation study on various classification-based unary losses is performed to show the effectiveness of the Semantic Cluster Unary Loss(SCUL). Sensitivity of parameters is also discussed in this section.

\subsection{Datasets and Evaluation Metrics}

In this section, we run large-scale retrieval experiments on three image benchmarks: CIFAR-10\footnote{http://www.cs.toronto.edu/\textasciitilde kriz/cifar.html}, Nuswide\footnote{http://lms.comp.nus.edu.sg/research/NUS-WIDE.htm} and ImageNet\footnote{http://image-net.org}. CIFAR-10 consists of 60,000 $32 \times 32$ color images from 10 object categories. ImageNet dataset is obtained from ILSVRC2012 dataset, which contains more than 1.2 million training images of 1,000 categories in total, together with 50,000 validation images. Nuswide dataset contains about 270K images collected from Flickr, and about 220K images are available from the Internet now. It associates with 81 ground truth concept labels, and each image contains multiple semantic labels. Following~\cite{liu2011hashing}, we only use the images associated with the 21 most frequent concept tags, where the total number of images is about 190K, and the number of images associated with each tag is at least 5,000.

The experimental protocols are similar to~\cite{xia2014supervised}. In CIFAR-10 dataset, we randomly select 1,000 images (100 images per class) as the query set, and the rest 59,000 images as the retrieval database. In ImageNet dataset, the provided training set are used for retrieval database, and 50,000 validation images for the query set. In Nuswide dataset, we randomly select 2,100 images (100 images per class) as the query set. For CIFAR-10 and ImageNet, similar data pairs share the same semantic label. For Nuswide dataset, similar images share at least one semantic label.

Our method is implemented with PyTorch\footnote{http://pytorch.org/} framework. We use pre-trained network parameters before $\mathrm{fc7}$ if necessary, and the parameters after $\mathrm{fc7}$ is initialized by "Gaussian" initializer with zero mean and standard deviation 0.01, except the Semantic Clusters $\mathbf{C}$ which are initialized with standard deviation 0.5 to make the norms of clusters larger. The images are resized to the proper input sizes to train the network (e.g. $224 \times 224$ for AlexNet and VGGNet). SGD is used for optimization, the momentum is 0.9 and the initial learning rate is set to 0.001 before $\mathrm{fc7}$ and 0.01 for the rest of layers. 

For training the deep hashing network, we randomly select 5,000 images (500 per class) in CIFAR-10 and 10,500 images (500 per class) in Nuswide to train the network, and use all database images in the ImageNet dataset for training. For semi-supervised hashing methods, the remaining database data are regraded as unlabeled samples. The hyper-parameters $\lambda, \mu, \alpha$ is different according to datasets, which are selected with the validation set. We first of all randomly select part of training data as the validation set to determine the parameters. For CIFAR-10, we choose $\{ \lambda=0.005, \mu=0.2, \alpha=0.05 \}$ and the learning rate decreases by 80\% after 100,140 epochs and stops at 160 epochs. For Nuswide, we select $\{ \lambda=0.001, \mu=0.1, \alpha=1.0 \}$ and the learning rate decreases by 80\% after 40 epochs and stops at 60 epochs. For ImageNet we select $\{ \lambda=0.001, \mu=0.1, \alpha=4.0 \}$ and the learning rate decreases by 80\% after 12, 17 epochs and stops at 20 epochs. Detailed parameter selection strategies are discussed in Sec. \ref{sec:param}. Unless specified, we just use the {\em warm-up} procedure for training ImageNet in which the norms of Semantic Cluster centers $\mathbf{C}$ are constrained to $s=8$ in the first 5 epochs. The training is done on a server with two Intel(R) Xeon(R) E5-2683 v3@2.0GHz CPUs, 256GB RAM and a Geforce GTX TITAN Pascal with 12GB memory.

Similar to~\cite{liu2012supervised,xia2014supervised}, for each retrieval dataset, we report the compared results in terms of {\em mean average precision}(MAP), precision at Hamming distance within 2, and precision of top returned candidates. For Nuswide, we calculate the MAP value within the top 5000 returned neighbors, and we report the MAP of all retrieved samples on CIFAR-10. Groundtruths are defined by whether two candidates are similar. We run each experiment for 5 times and get the average result.

\begin{table}[t]
    \setlength{\abovecaptionskip}{3pt}
    \setlength{\belowcaptionskip}{0pt}
    \centering
    \footnotesize
    \begin{tabular}{c|c|cccc}
        \hline
         & & \multicolumn{4}{c}{MAP} \\
        Method & Net & 12 bits & 24 bits & 32 bits & 48 bits \\
        \hline
        \multicolumn{6}{c}{Fine-tuning from AlexNet or VGG-F Net} \\
        \hline
        DHN~\cite{zhu2016deep} & AlexNet & 0.555 & 0.594 & 0.603 & 0.621 \\
        DQN~\cite{cao2016deep} & AlexNet & 0.554 & 0.558 & 0.564 & 0.580 \\
        DPSH~\cite{li2015feature} & VGG-F & 0.682 & 0.686 & 0.725 & 0.733 \\
        DSH~\cite{yansemi}* & VGG-F & 0.604 & 0.746 & 0.781 & 0.810 \\
        DISH~\cite{zhang2017scalable} & AlexNet & 0.758 & 0.784 & 0.799 & 0.791 \\
        DSDH~\cite{li2017deep} & VGG-F & 0.740 & 0.786 & 0.801 & 0.820 \\
        CNNBH* & AlexNet & 0.794 & 0.809 & 0.808 & 0.814 \\
        SSDH* & AlexNet & 0.789 & 0.808 & 0.813 & 0.822 \\
        Cls-LSH* & AlexNet & 0.749 & 0.766 & 0.778 & 0.780 \\
        Cls-onehot* & AlexNet & \multicolumn{4}{c}{0.715} \\
        \textbf{SCDH(Ours)} & AlexNet & \textbf{0.801} & \textbf{0.822} & \textbf{0.828} & \textbf{0.836} \\
        \hline
        \multicolumn{6}{c}{Fine-tuning from VGG-16 Net} \\
        \hline
        NINH~\cite{zhuang2016fast}* & VGG-16 & N/A & 0.677 & 0.688 & 0.699 \\
        FTDE~\cite{zhuang2016fast} & VGG-16 & N/A & 0.760 & 0.768 & 0.769 \\
        BOH~\cite{dai2016binary} & VGG-16 & 0.620 & 0.633 & 0.644 & 0.657 \\
        DSH* & VGG-16 & 0.818 & 0.825 & 0.847 & 0.849 \\
        DISH~\cite{zhang2017scalable} & VGG-16 & 0.841 & 0.854 & 0.859 & 0.857 \\
        DRLIH~\cite{zhang2018deep} & VGG-16 & 0.816 & 0.843 & 0.855 & 0.853\\
        SSDH* & VGG-16 & \textbf{0.845} & 0.856 & 0.863 & 0.865 \\
        CNNBH* & VGG-16 & 0.833 & 0.847 & 0.855 & 0.855 \\
        \textbf{SCDH(Ours)} & VGG-16 & 0.841 & \textbf{0.860} & \textbf{0.865} & \textbf{0.870} \\
        \hline
        \multicolumn{6}{c}{Fine-tuning from ResNet-50 Net} \\
        \hline
        DSH* & ResNet-50 & 0.886 & 0.900 & 0.896 & 0.895 \\
        CNNBH* & ResNet-50 & \textbf{0.895} & 0.902 & 0.910 & 0.905 \\
        SSDH* & ResNet-50 & 0.894 & \textbf{0.910} & 0.911 & 0.915 \\
        \textbf{SCDH(Ours)} & ResNet-50 & 0.894 & \textbf{0.910} & \textbf{0.913} & \textbf{0.918} \\
        \hline
    \end{tabular}
    \caption{Results of deep hashing methods in MAP on CIFAR-10. For this dataset, 5,000 data are randomly sampled as training set. The classification performance of AlexNet is slightly worse than VGG-F. * denotes re-running the code in the corresponding papers or our reimplementation. The results of the proposed SCDH are the average of 5 trails.}
    \label{tab:cifar10_deep}
\end{table}

\begin{table}[t]
    \setlength{\abovecaptionskip}{3pt}
    \setlength{\belowcaptionskip}{0pt}
    \centering
    \footnotesize
    \begin{tabular}{c|c|cccc}
	    \hline
         & & \multicolumn{4}{c}{MAP} \\
        Method & Net & 12 bits & 24 bits & 32 bits & 48 bits \\
        \hline
        \multicolumn{6}{c}{Fine-tuning from AlexNet or VGG-F Net} \\
        \hline
        DHN~\cite{zhu2016deep} & AlexNet & 0.708 & 0.735 & 0.748 & 0.758 \\
        DQN~\cite{cao2016deep} & AlexNet & 0.768 & 0.776 & 0.783 & 0.792 \\
        DPSH~\cite{li2015feature} & VGG-F & 0.794 & 0.822 & 0.833 & \textbf{0.851} \\
        DSH~\cite{yansemi}* & VGG-F & 0.751 & 0.765 & 0.767 & 0.773 \\
        DISH~\cite{zhang2017scalable} & AlexNet & 0.787 & 0.810 & 0.810 & 0.813 \\
        DSDH~\cite{li2017deep} & VGG-F & 0.776 & 0.808 & 0.820 & 0.829 \\
        SSDH* & AlexNet & 0.775 & 0.796 & 0.800 & 0.807 \\
        \textbf{SCDH(Ours)} & AlexNet & \textbf{0.804} & \textbf{0.834} & \textbf{0.842} & 0.850 \\
        \hline
        \multicolumn{6}{c}{Fine-tuning from VGG-16 Net} \\
        \hline
        NINH~\cite{zhuang2016fast}* & VGG-16 & N/A & 0.718 & 0.720 & 0.723 \\
        FTDE~\cite{zhuang2016fast} & VGG-16 & N/A & 0.750 & 0.756 & 0.760 \\
        BOH~\cite{dai2016binary} & VGG-16 & 0.786 & 0.834 & 0.837 & 0.855 \\
        DISH~\cite{li2017deep} & VGG-16 & 0.833 & 0.850 & 0.850 & 0.856 \\
        DRLIH~\cite{zhang2018deep} & VGG-16 & 0.823 & 0.846 & 0.845 & 0.853 \\
        SSDH* & VGG-16 & 0.820 & 0.840 & 0.845 & 0.848 \\
        \textbf{SCDH(Ours)} & VGG-16 & \textbf{0.836} & \textbf{0.857} & \textbf{0.860} & \textbf{0.868} \\
        \hline
        \multicolumn{6}{c}{Fine-tuning from ResNet-50 Net} \\
        \hline
        SSDH* & ResNet-50 & 0.794 & 0.815 & 0.810 & 0.816 \\
        \textbf{SCDH(Ours)} & ResNet-50 & \textbf{0.836} & \textbf{0.868} & \textbf{0.872} & \textbf{0.878} \\
        \hline
    \end{tabular}
    \caption{Results of deep hashing methods in MAP on Nuswide dataset. * denotes re-running the code in the corresponding papers or our own implementation. The results of the proposed SCDH are the average of 5 trails.}
    \label{tab:nuswide_deep}
\end{table}

\begin{table}[t]
    \setlength{\abovecaptionskip}{3pt}
    \setlength{\belowcaptionskip}{0pt}
    \centering
    \footnotesize
    \begin{tabular}{c|c|c}
	    \hline
        Method & Net &  MAP(128 bits) \\
        \hline
        SDH~\cite{Shen_2015_CVPR}* & VGG-19 & 0.313 \\
        DISH~\cite{zhang2017scalable} & VGG-19 & 0.452 \\
        \textbf{SCDH(Ours)} & AlexNet & 0.441 \\
        \textbf{SCDH(Ours)} & VGG-19 & \textbf{0.603} \\
        \textbf{SCDH(Ours)} & ResNet-152 & \textbf{0.694} \\
        \hline
    \end{tabular}
    \caption{Results of various hashing methods in MAP on ImageNet dataset. * denotes re-running the code in the corresponding papers. We use $\mathrm{fc7}$ features in VGG-19 net for training SDH~\cite{lin2015supervised}.}
    \label{tab:imagenet_deep}
\end{table}

\subsection{Comparison on Supervised Hashing}

\begin{table}[t]
    \setlength{\abovecaptionskip}{3pt}
    \setlength{\belowcaptionskip}{0pt}
    \centering
    \footnotesize
    \begin{tabular}{c|c|cc}
        \hline
         & & \multicolumn{2}{c}{Training time(hours)} \\
        Method & Net & CIFAR-10 & Nuswide \\
        \hline
        NINH~\cite{zhuang2016fast}* & VGG-16 & 174 & 365 \\
        FTDE~\cite{zhuang2016fast} & VGG-16 & 15 & 32 \\
        DISH & VGG-16 & 4 & 9 \\
        DSH & VGG-16 & 3 & 5 \\
        \textbf{SCDH(Ours)} & VGG-16 & \textbf{0.9} & \textbf{0.7} \\
        \hline
    \end{tabular}
    \caption{Training time(in hours) of various deep hashing methods. VGG-16 net is used for evaluation. The codes of NINH is re-runned by ~\cite{zhuang2016fast}.}
    \label{tab:deep_time}
\end{table}

We compare our SCDH method with recent state-of-the-art deep hashing methods, including pairwise based methods such as DSH~\cite{liu2016deep}, DHN~\cite{zhu2016deep}, DPSH~\cite{li2015feature}, DQN~\cite{cao2016deep}, DISH~\cite{zhang2017scalable}, DSDH~\cite{li2017deep}, triplet based methods like NINH~\cite{lai2015simultaneous}, FTDE~\cite{zhuang2016fast}, BOH~\cite{dai2016binary}, DRLIH~\cite{zhang2018deep}, and unary loss based methods like CNNBH~\cite{guo2016hash}, SSDH~\cite{yang2018supervised}. They follow similar experimental settings, but different methods may use different deep networks, thus we train on several types of network (AlexNet, VGGNet, ResNet, etc.) for fair comparison.

\begin{figure}[t]
    \setlength{\abovecaptionskip}{1pt}
    \setlength{\belowcaptionskip}{0pt}
    \centering
    \includegraphics[scale=0.242]{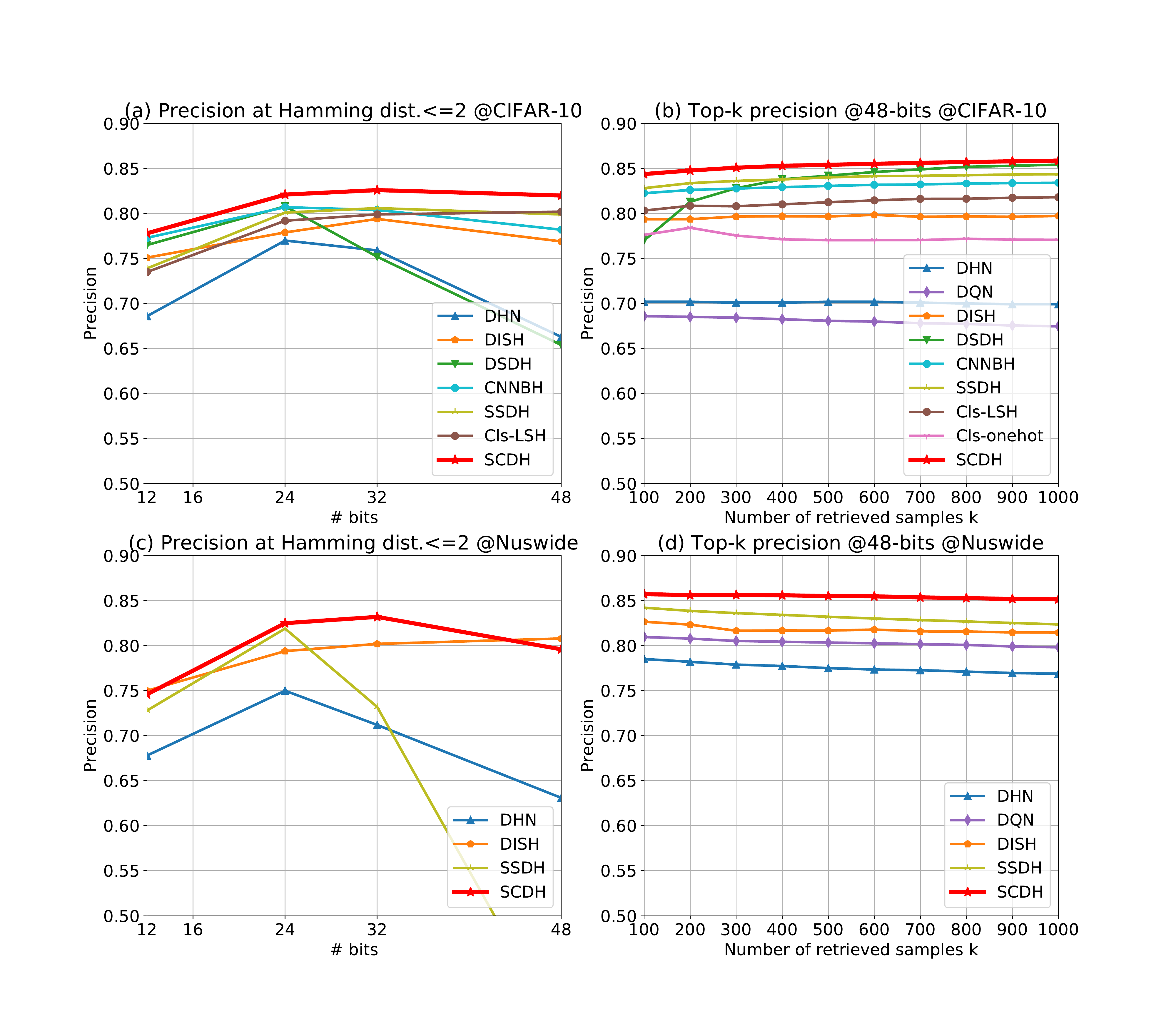}
    \caption{Precision at Hamming distance within 2 value and top-k precision curve of different deep hashing methods on CIFAR-10 and Nuswide dataset. AlexNet is used for pre-training in these algorithms.}
    \label{fig:deephash}
\end{figure}

Retrieval results of different methods are shown in Table \ref{tab:cifar10_deep},\ref{tab:nuswide_deep},\ref{tab:imagenet_deep} and Figure \ref{fig:deephash}. Note that the results with citations are copied from the corresponding papers, and some results are our own implementation with suggested parameters of the original paper. With the network structure fixed, our SCDH algorithm achieves better performance than methods with pairwise or triplet losses like NINH, DHN, DPSH, DSH and BOH, especially on CIFAR-10, showing the effectiveness of the proposed SCUL. As the discrete hashing methods like DISH and DSDH cannot be trained directly by back-propagation due to the special consideration of the discrete constraints, our SCDH method performs better than those methods. Moreover, our method performs better than the unary loss based methods like CNNBH and SSDH on most settings, showing that the proposed SCUL is friendly with the distance learning. It should be noticed that the classification performance of VGG-F net is slightly better than AlexNet, thus the hashing performance is expected not to decrease and may even get better if replacing AlexNet with VGG-F. 

As the similarity information is defined by the semantic labels, it seems more simple to directly convert the predicted label to binary codes. As discussed in~\cite{sablayrolles2017should}, we can encode the predicted label with one-hot binary codes, or convert the class probability to binary codes using LSH. We name these two methods as Cls-onehot and Cls-LSH respectively. Results on CIFAR-10 dataset are shown in Table \ref{tab:cifar10_deep} and Figure \ref{fig:deephash}. The performances are lower than many hashing methods. Similar conclusion is arrived at~\cite{wan2014deep}, which shows that latent features extracted from fc6/fc7 contain more semantic information than the class probability features. Moreover, one-hot encodings lose more information than the class probabilities, thus Cls-onehot performs inferior than Cls-LSH. 

Table \ref{tab:deep_time} summarizes the training time of some state-of-the-art methods. VGG-16 net is used for evaluation. As expected, the training speed of the proposed method is much faster than most deep hashing methods with pairwise losses or triplet losses. It takes less than 1 hour to generate good binary codes by the VGG-16 net, thus we can also train binary codes efficiently with deep neural nets.

\begin{table}[t]
    \setlength{\abovecaptionskip}{0pt}
    \setlength{\belowcaptionskip}{0pt}
    \centering
    \footnotesize
    \begin{tabular}{c|c|cccc}
	    \hline
         & & \multicolumn{4}{c}{MAP} \\
        Method & Net & 12 bits & 24 bits & 32 bits & 48 bits \\
        \hline
        \multicolumn{5}{c}{Hashing on CIFAR-10 Dataset} \\
        \hline
        SemiSDH~\cite{zhang2017ssdh} & VGG-F & 0.801 & 0.813 & 0.812 & 0.814 \\
        BGDH~\cite{yan2017semi} & VGG-F & 0.805 & 0.824 & 0.826 & 0.833 \\
        \textbf{MT-SCDH(Ours)} & AlexNet & \textbf{0.828} & \textbf{0.844} & \textbf{0.849} & \textbf{0.855} \\
        \hline
        \multicolumn{5}{c}{Hashing on Nuswide Dataset} \\
        \hline
        SemiSDH~\cite{zhang2017ssdh} & VGG-F & 0.773 & 0.779 & 0.778 & 0.778 \\
        BGDH~\cite{yan2017semi} & VGG-F & 0.803 & 0.818 & 0.822 & 0.828 \\
        \textbf{MT-SCDH(Ours)} & AlexNet & \textbf{0.811} & \textbf{0.838} & \textbf{0.843} & \textbf{0.853} \\
        \hline
    \end{tabular}
    \caption{Results of the semi-supervised hashing methods on CIFAR-10 and Nuswide dataset. AlexNet is used for pre-training. Note that we use 5,000 labeled images in CIFAR-10 and 10,500 labeled images in the Nuswide dataset, and regard the rest as unlabeled images.}
    \label{tab:ssh_deep}
\end{table}

\subsection{Comparison on Semi-Supervised Hashing}

We compare our MT-SCDH method with recent state-of-the-art semi-supervised deep hashing methods including SemiSDH~\cite{zhang2017ssdh}, BGDH~\cite{yan2017semi}. AlexNet is used for fair comparison. As suggested in~\cite{tarvainen2017mean}, we use $w=50$ in the experiments. Retrieval results are shown in Table~\ref{tab:ssh_deep}. It is clear that the MT-SCDH algorithm performs much better than others by over 2 percents. Compared with the results of SCDH shown in Table \ref{tab:cifar10_deep} and \ref{tab:nuswide_deep}, the semi-supervised setting achieves better MAP value by about 0.3-2 percents, showing that the Mean Teacher based semi-supervised hashing approach is able to capture more semantic information with the unlabeled data.

\subsection{Ablation Study}

\begin{table}[t]
    \setlength{\abovecaptionskip}{0pt}
    \setlength{\belowcaptionskip}{0pt}
    \centering
    \footnotesize
    \begin{tabular}{c|cccc|cc}
	    \hline
         & \multicolumn{4}{c}{MAP} & \multicolumn{2}{|c}{Precision} \\
        Method & 12 bits & 24 bits & 32 bits & 48 bits & 32 bits & 48 bits \\
        \hline
        \multicolumn{7}{c}{Hashing on CIFAR-10 Dataset} \\
        \hline
        SCDH-S & 0.753 & 0.788 & 0.797 & 0.810 & 0.806 & 0.786 \\
        SCDH-C & 0.789 & 0.810 & 0.822 & 0.829 & 0.819 & 0.799 \\
        SCDH & \textbf{0.801} & \textbf{0.822} & \textbf{0.828} & \textbf{0.836} & \textbf{0.826} & \textbf{0.820} \\
        \hline
        \multicolumn{7}{c}{Hashing on Nuswide Dataset} \\
        \hline
        SCDH-S & 0.739 & 0.779 & 0.788 & 0.804 & 0.800 & 0.736 \\
        SCDH-C & 0.744 & 0.778 & 0.790 & 0.803 & 0.798 & 0.791 \\
        SCDH & \textbf{0.804} & \textbf{0.834} & \textbf{0.842} & \textbf{0.850} & \textbf{0.832} & \textbf{0.796} \\
        \hline
        \multicolumn{7}{c}{Hashing on ImageNet Dataset (48 bits)} \\
        \hline
        SCDH-S & \multicolumn{4}{c|}{0.337} & \multicolumn{2}{c}{0.282} \\
        SCDH-C & \multicolumn{4}{c|}{0.334} & \multicolumn{2}{c}{0.298} \\
        SCDH & \multicolumn{4}{c|}{\textbf{0.421}} & \multicolumn{2}{c}{\textbf{0.344}} \\
        \hline
    \end{tabular}
    \caption{Results of the variants of the proposed SCDH algorithm on CIFAR-10, Nuswide and ImageNet dataset. AlexNet is used for pre-training. Precision denotes the precision at Hamming distance within 2 value.}
    \label{tab:variant_deep}
\end{table}

\begin{figure}[t]
    \setlength{\abovecaptionskip}{0pt}
    \setlength{\belowcaptionskip}{0pt}
    \centering
    \includegraphics[scale=0.242]{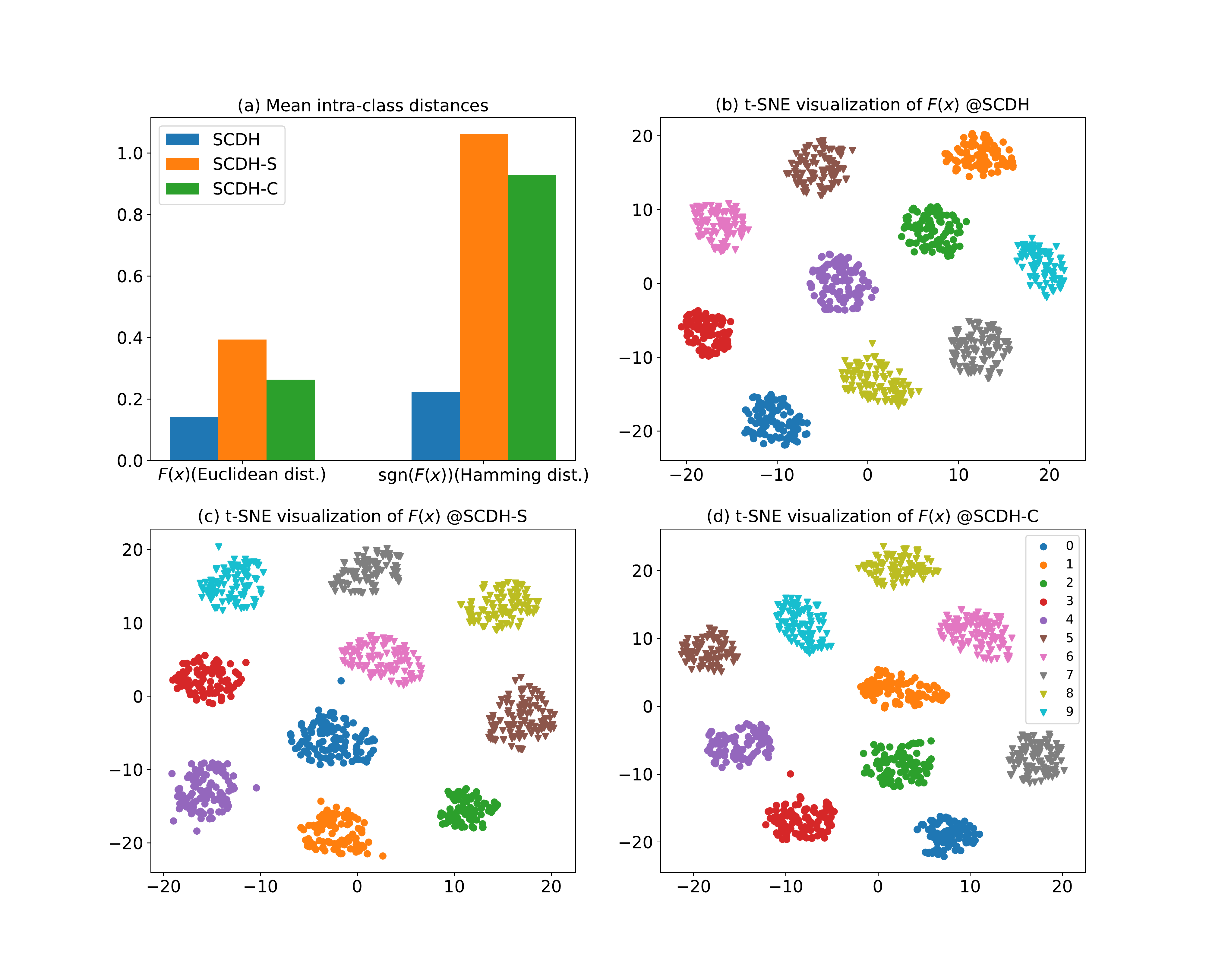}
    \caption{Visualization on the real-value representations $F(\mathbf{x})$ and hashcodes $\mathrm{sgn}(F(\mathbf{x}))$ learned by SCDH and its variants. the code length is 48.}
    \label{fig:illus-vari}
\end{figure}

\begin{table}[t]
    \setlength{\abovecaptionskip}{0pt}
    \setlength{\belowcaptionskip}{0pt}
    \centering
    \footnotesize
    \begin{tabular}{c|c|c|c}
	    \hline
        Initialization & MAP & Precision & Ave. Norm of SCC. \\
        \hline
        Random ($\sigma=0.5$) & 0.385 & 0.338 & 5.41 \\
        {\em warm-up} ($s=8$) & 0.419 & \textbf{0.348} & 5.75 \\
        {\em warm-up} ($s=4$) & \textbf{0.421} & 0.344 & 5.69 \\
        \hline
    \end{tabular}
    \caption{Results of whether to use {\em warm-up} initialization for training SCDH on ImageNet dataset. AlexNet is used for pre-training. Precision denotes the precision at Hamming distance within 2 value. Random denotes use gaussian initialization with standard deviation 0.5. {\em Warm-up} denotes using {\em warm-up} procedure where the norm of centers $s$ is constrained to 4 and 8 respectively. The average norm of SCC (Semantic Cluster centers) after optimization is also provided. The code length is 48.}
    \label{tab:variant_warm}
\end{table}

\textbf{Variants of SCDH} In order to verify the effectiveness of our method, several variants of the proposed method are also considered. First, we regard $\mathcal{L}_u$ as just softmax loss such that $\lambda = 0, l_c(\mathbf{h}_i, y_i) = - \log \frac{\exp \mathbf{c}_{y_i}^\mathrm{T} \mathbf{h}_i}{\sum_{l=1}^C \exp \mathbf{c}_l^\mathrm{T} \mathbf{h}_i} $. It is very similar with CNNBH~\cite{guo2016hash} and we name it as SCDH-S. Second, we apply CenterLoss~\cite{wen2016discriminative} for hashing, denote SCDH-C, in which $\mu=0$ and we replace $\mathcal{L}_u$ with the softmax loss and centerloss defined in~\cite{wen2016discriminative}. Inspired by the SCUL in multilabel case, we extend the CenterLoss to the multilabel case for SCDH-C such that
\begin{equation}
\mathcal{L} = \sum_{i=1}^n \frac{1}{|Y_i|}[\sum_{s \in Y_i} [ -\log \frac{\exp \mathbf{c}_{s}^\mathrm{T} \mathbf{h}_i}{\sum_{j=1}^C \exp\mathbf{c}_{j}^\mathrm{T} \mathbf{h}_i} ] + \lambda \sum_{s \in Y_i} \|\mathbf{h}_i - \mathbf{w}_s\|^2]
\end{equation}

Retrieval results are shown in Table \ref{tab:variant_deep}. It is clear that SCDH and its variants perform better than most deep hashing algorithms. In particular, the proposed SCDH algorithm performs better than the variants, especially on the precision at Hamming distance within 2 value. Moreover, SCDH performs much better on Nuswide dataset, showing the effectiveness of the SCUL over the modified CenterLoss and the softmax loss on the multilabel version.

To address the above issue, Figure \ref{fig:illus-vari}(a) shows the average intra-class distances of the normalized real-value representations $F(\mathbf{x})$ and hashcodes $\mathrm{sgn}(F(\mathbf{x}))$ of the training set. It is clear that the intra-class distances of the representations learned by SCDH is much smaller than those learned by the variants, thus SCDH is more likely to hash more similar data to the same hashcode, improving the precision at Hamming distance within 2 value. Figure \ref{fig:illus-vari}(b-d) are t-SNE visualizations of the normalized $F(\mathbf{x})$ trained by SCDH and its variants. The clusters learned by SCDH is more compact than the variants, and there are less outliers in SCDH. Furthermore, some outliers in SCDH-S and SCDH-C goes to other clusters (like label 0 in SCDH-S and label 3 in SCDH-C). To conclude, the proposed SCUL have close relationship with the triplet ranking loss and get better results than CenterLoss and softmax loss, especially on the multilabel case.

\textbf{The {\em warm-up} training procedure} Another important issue is whether to use the {\em warm-up} procedure to optimize $\mathbf{C}$ for training datasets with large amount of labels. Table \ref{tab:variant_warm} shows results on ImageNet dataset, which implies that the {\em warm-up} procedure performs better than the random initialization procedure. The performance of the {\em warm-up} procedure with different norms of clusters $s$ performs almost the same, including the average norm of the Semantic cluster centers after optimization, which implies that the results is not sensitive to $s$ in the {\em warm-up} procedure, thus we can set this parameter freely for training large datasets.

\subsection{Sensitivity to Parameters}
\label{sec:param}

\begin{figure}[t]
    \setlength{\abovecaptionskip}{0pt}
    \setlength{\belowcaptionskip}{0pt}
    \centering
    \includegraphics[scale=0.235]{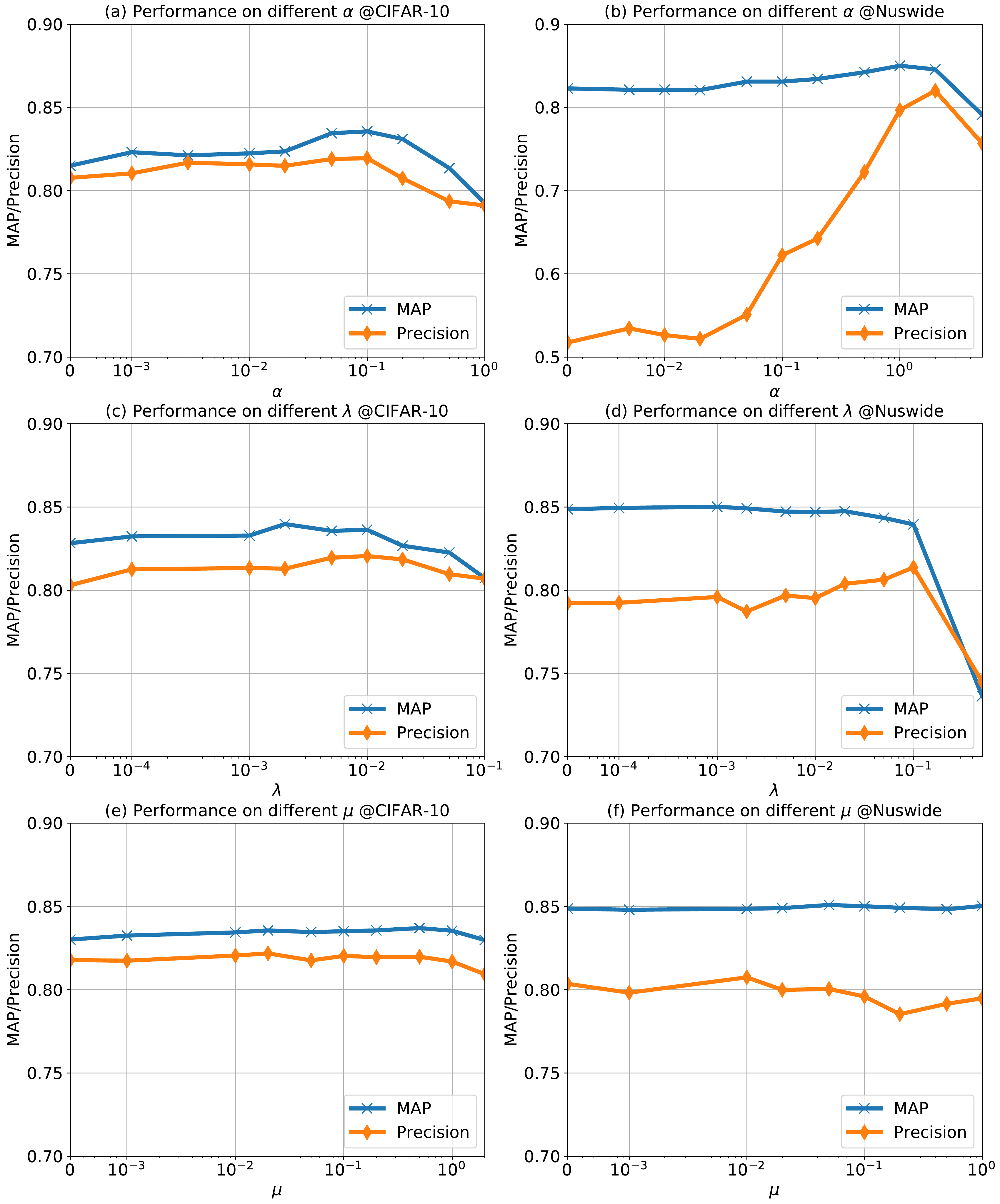}
    \caption{Comparative results of different hyper-parameters on CIFAR-10 and Nuswide dataset. The code length is 48.}
    \label{fig:param}
\end{figure}

\begin{figure}[t]
    \setlength{\abovecaptionskip}{0pt}
    \setlength{\belowcaptionskip}{0pt}
    \centering
    \includegraphics[scale=0.235]{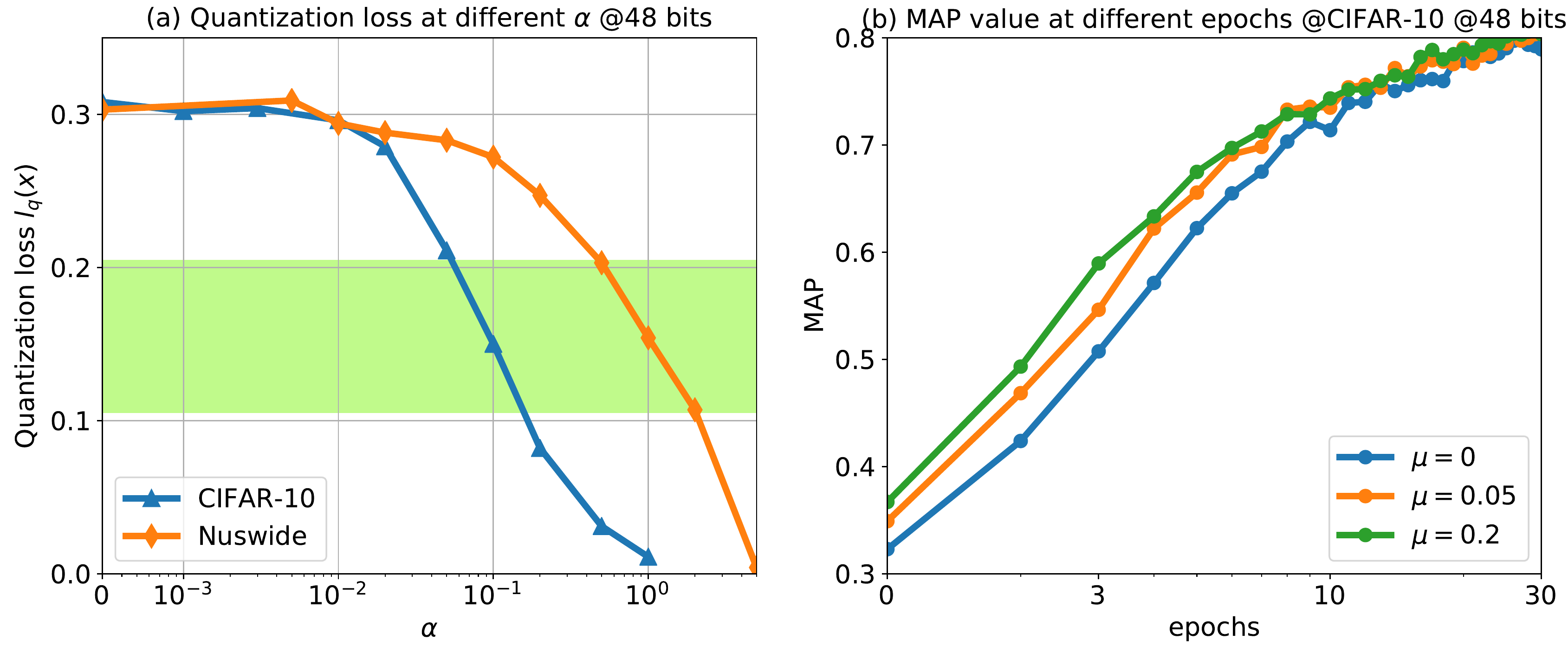}
    \caption{Detailed analysis on $\alpha$ and $\mu$. The light green background in (a) denotes the proper region for setting $\alpha$.}
    \label{fig:analysis}
\end{figure}

In this section, influence on different settings of the proposed SCDH algorithm is evaluated. We use AlexNet for pre-training, and the code length is 48.

\textbf{Influence of $\alpha$} Figure \ref{fig:param}(a)(b) shows the performance on different values of $\alpha$. It can be seen clearly that setting a certain $\alpha$ achieves better performance on either MAP or precision at Hamming distance within 2 value than setting $\alpha=0$. It means that adding the quantization loss improves the performance of hashing. 

To get better hashcodes, a proper value of $\alpha$ should be set. Figure \ref{fig:analysis}(a) shows the quantization loss at the end of training with different $\alpha$. Compared with \ref{fig:param}(a)(b), $\alpha$ should be set to get the best performance in that the training quantization loss should be around $0.1 \sim 0.2$.

\textbf{Influence of $\lambda$} Figure \ref{fig:param}(c)(d) shows the performance on different values of $\lambda$. It shows that a relatively small $\lambda$ leads to better performance on MAP. In fact, only optimizing $l_c(\mathbf{h}_i, y_i)$ is able to minimize the intra-class distances. Although $\lambda$ is not effective to get better MAP, greater value of precision at Hamming distance with 2 may be achieved when $\lambda$ goes larger, especially on the Nuswide dataset. It implies that the term $|\mathbf{h}_i - \mathbf{c}_{y_i}|$ is an auxiliary term to make intra-class smaller, thus more similar data is expected to map to the same hashcode.

\textbf{Influence of $\mu$} Figure \ref{fig:param}(e)(f) shows the performance on different values of $\mu$. It is shown that the algorithm is not too sensitive to $\mu$ over a wide range, but the performance is slightly better when $\mu \sim 0.1$, thus we recommend to set $\mu \sim 0.1$.

Figure \ref{fig:analysis}(b) shows the MAP value with different training epochs on CIFAR-10 dataset. It can be seen clearly that a larger $\mu$ makes the training procedure faster, showing that setting proper $\mu$ is helpful for faster convergence.

\section{Conclusion and Future Work}
\label{sec:conclusion}

In this paper, we propose a novel and efficient supervised hashing algorithm. We first of all introduce a Unary Upper Bound of the traditional triplet loss, thus bridging the triplet loss and the classification-based unary loss. The Unary Upper Bound shows that each semantic label corresponds to a certain cluster in the Hamming space, and minimizing the Unary Upper Bound is expected to minimize the intra-class distances and separate different clusters far apart, thus the traditional triplet loss are minimized. Second, we propose a novel supervised hashing algorithm named {\em Semantic Cluster Deep Hashing (SCDH)}, in which the loss is the modified Unary Upper Bound, named Semantic Cluster Unary Loss(SCUL). Third, we extend the SCDH algorithm to semi-supervised hashing by combining the state-of-the-art Mean Teacher algorithm with the SCUL. Experimental results on several supervised hashing datasets demonstrate the effectiveness of the proposed hashing algorithm.

Despite the success of the proposed SCDH algorithm, it should be noticed that our method is based on the semantic labels or tags. A certain unary loss for hashing with only pairwise similarity information should be discovered.


{\small
\bibliographystyle{ieee}
\bibliography{egbib}
}

\begin{IEEEbiography}[{\includegraphics[width=1in,height=1.25in,clip,keepaspectratio]{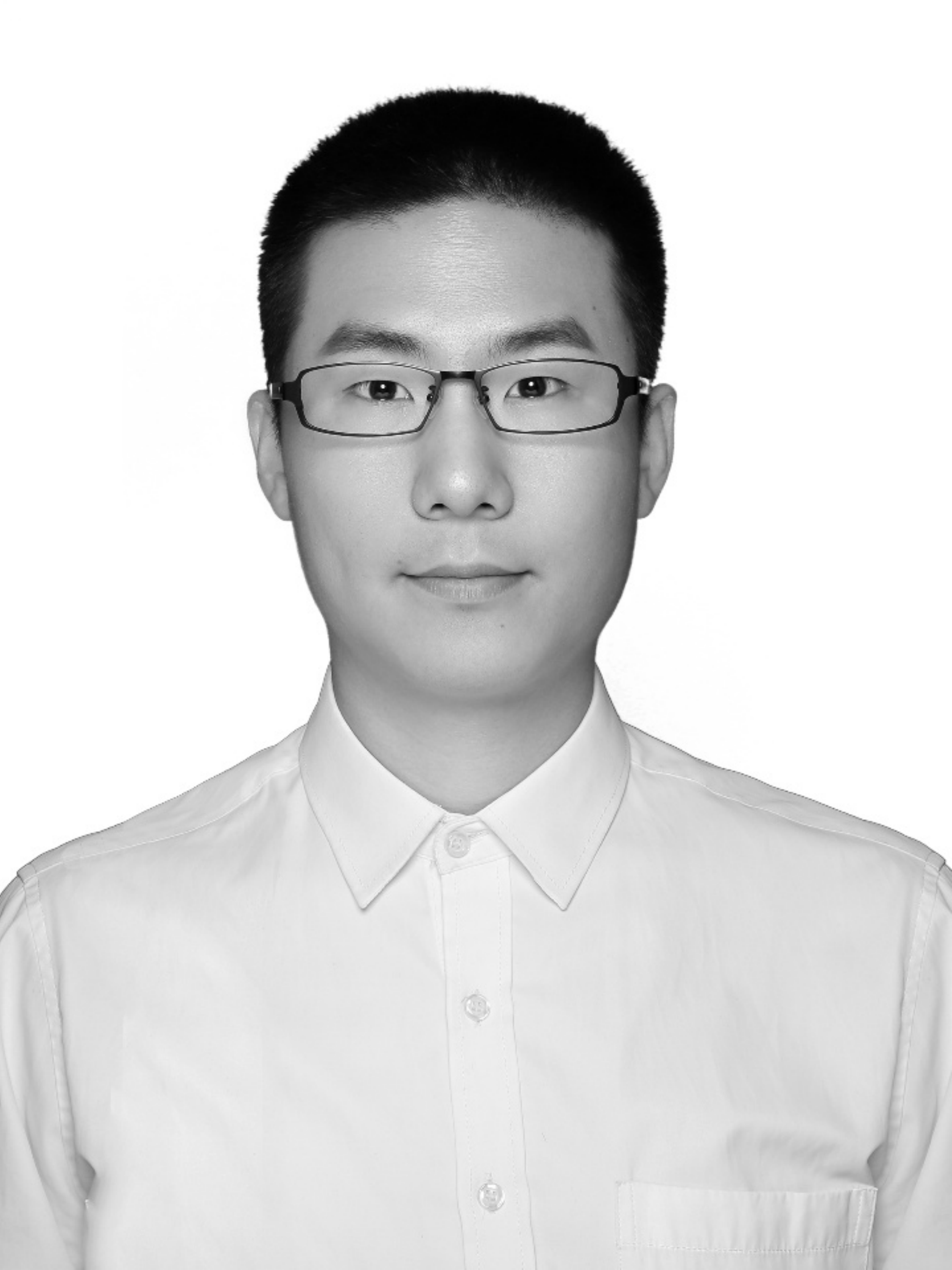}}]{Shifeng Zhang}
received the B.E. degree from the Department of Computer Science and Technology, Tsinghua University, in 2015. Now he is a Ph.D. student in the State Key Laboratory of Intelligent Technology and Systems, Beijing National Research Center for Information Science and Technology, Department of Computer Science and Technology, Tsinghua University. His
supervisor is Prof. Bo Zhang. His current research interests include hashing method and deep metric learning.
\end{IEEEbiography}

\begin{IEEEbiography}[{\includegraphics[width=1in,height=1.25in,clip,keepaspectratio]{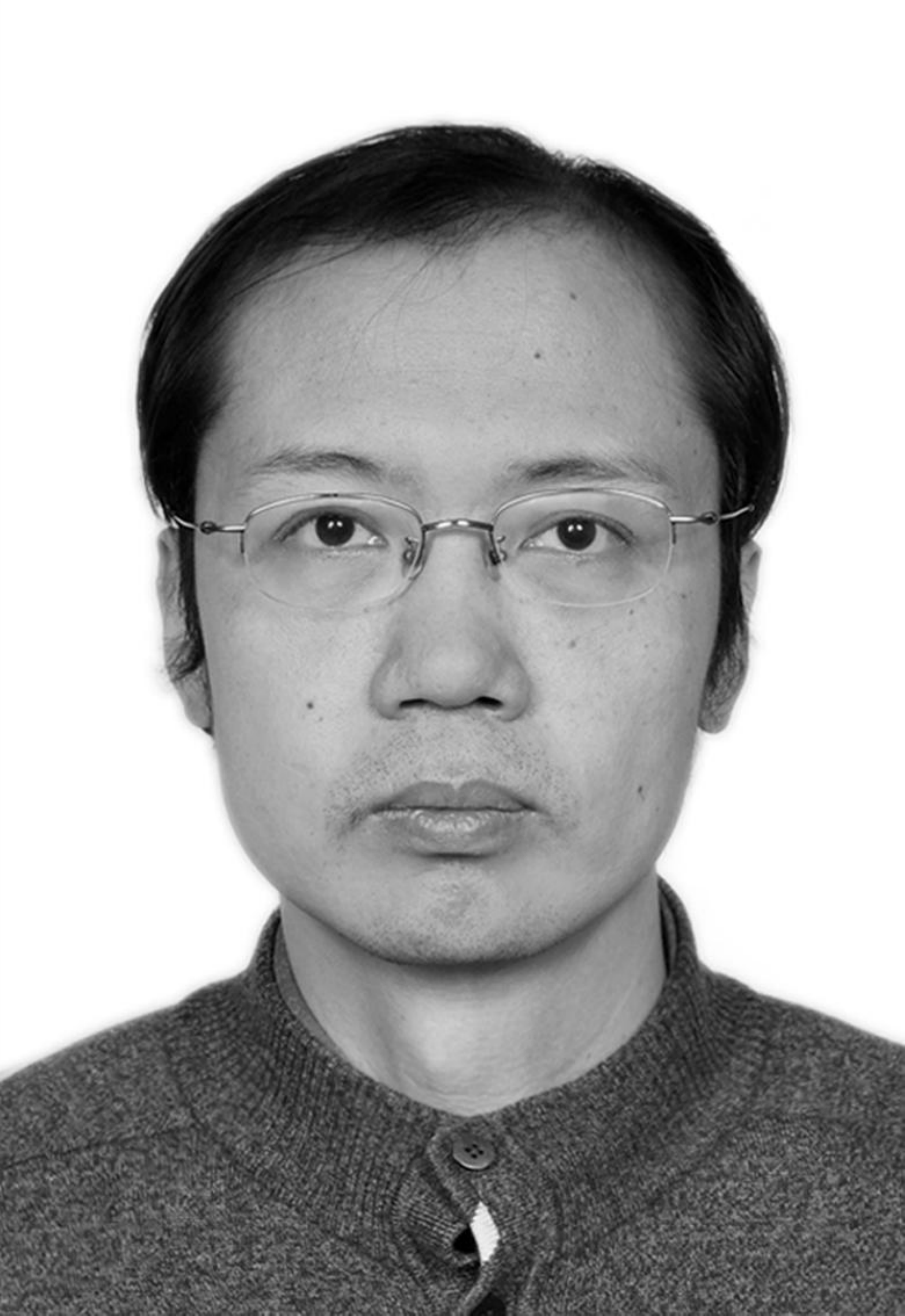}}]{Jianmin Li}
received the Ph.D. degree in computer application from the Department of Computer Science and Technology, Tsinghua University in 2003. Currently, he is an Associate Professor in the Department of Computer Science and Technology, Tsinghua University. His main research interests include image and video analysis, image and video retrieval and machine learning. He has published more than fifty journal and conference papers. He received the second class Technology Innovation Award by State Administration of Radio Film and Television in 2009.
\end{IEEEbiography}


\begin{IEEEbiography}[{\includegraphics[width=1in,height=1.25in,clip,keepaspectratio]{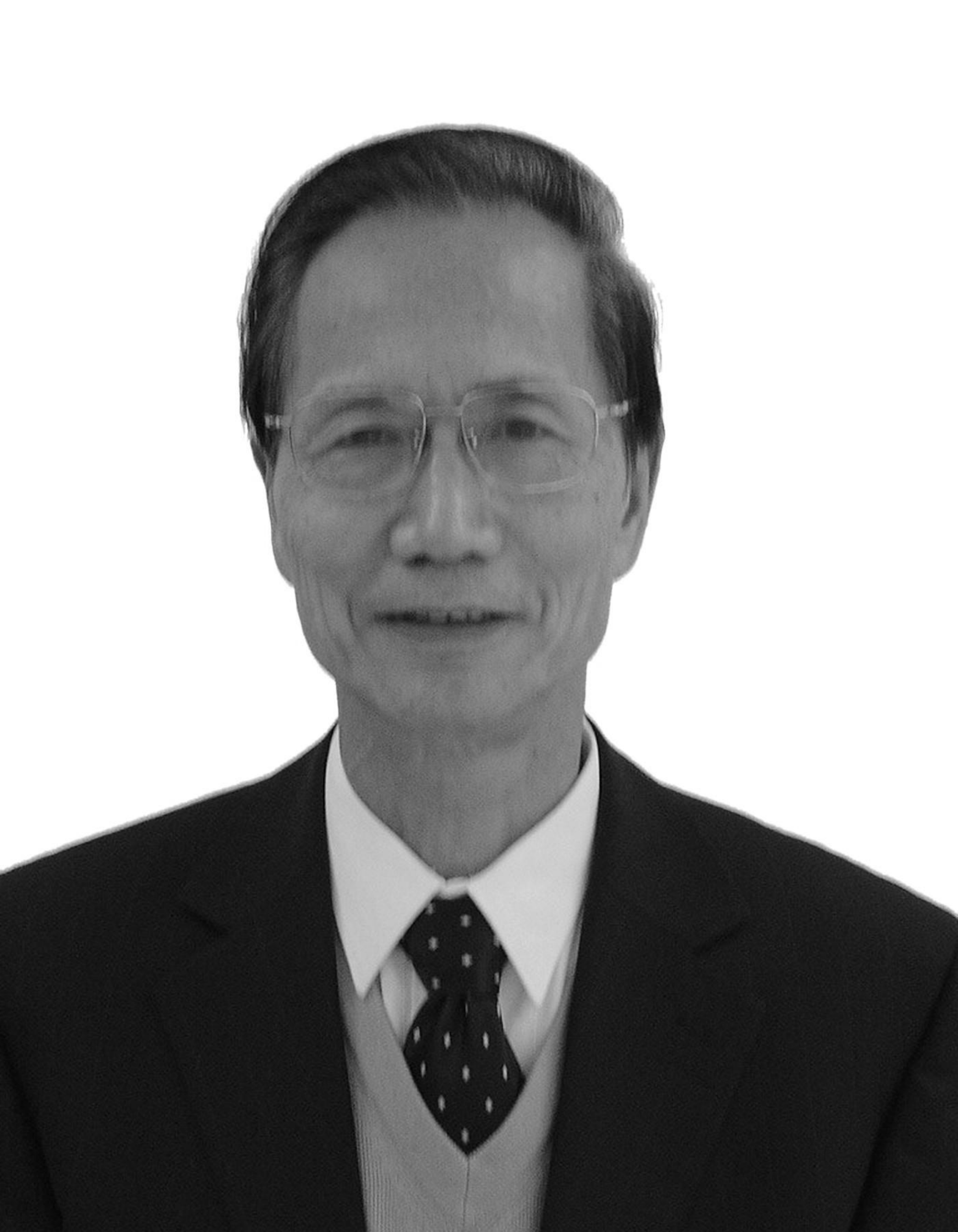}}]{Bo Zhang} 
graduated from Dept. of Automatic Control, Tsinghua University in 1958. He is now a professor of Computer Science and Technology Department, Tsinghua University, Beijing, China, the fellow of Chinese Academy of Sciences. His main research interests include artificial intelligence, robotics, intelligent control and pattern recognition. He has published about 150 papers and 3 monographs in these fields.
\end{IEEEbiography}

\onecolumn
\begin{appendices}

\section{Notations and Definitions}

The notations in the supplemental material are the same as that in the original paper, some of which are emphasized below.

Denote there are $n$ data instances $\mathbf{x}_1,...,\mathbf{x}_n$, each data instance $\mathbf{x}_i$ has a semantic label $y_i \in \{1,2,...,C \}$, and the hashcode of $\mathbf{x}_i$ is $\mathbf{h}_i$. Denote $S$ as the a set such that $(i, j) \in S$ implies $\mathbf{x}_i, \mathbf{x}_j$ are similar, and $(i, k) \notin S$ implies $\mathbf{x}_i, \mathbf{x}_k$ are dissimilar. Denote $g(\cdot, \cdot)$ as a monotonous, Lipschitz continuous function such that
\begin{equation}
\begin{split}
g(a,b) &\ge 0 \\
0 \le g(a_2, b) - g(a_1, b) \le a_2 - a_1, \quad a_1 &\le a_2 \\
0 \le g(a, b_1) - g(a, b_2) \le b_2 - b_1, \quad b_1 &\le b_2 \\
\end{split}
\label{eq:g}
\end{equation}

For the multiclass case, denote $y_i \in \{1,2,...,C \}, i = 1,2,..,n$ as the semantic label of the data instance $\mathbf{x}_i$. The data pairs are similar if they share the same semantic label, namely $y_i = y_j \Leftrightarrow (i, j) \in S$. The triplet ranking loss is defined as
\begin{equation}
\mathcal{L}_t = \sum_{(i,j) \in S, (i,k) \notin S}  g(|\mathbf{h}_i - \mathbf{h}_j|, |\mathbf{h}_i - \mathbf{h}_k|)
\label{eq:trf}
\end{equation}
where $g(\cdot, \cdot)$ has the property defined in Eq. (\ref{eq:g}). 

For the multilabel case, denote $Y_i \subseteq \{ 1,2,...,C \}$ as the labels of the data instance $\mathbf{x}_i$. The similarity is defined by whether $\mathbf{x}_i, \mathbf{x}_j$ share certain amount of semantic labels. The triplet ranking loss is defined as 
\begin{equation}
\mathcal{L}_{mt} = \sum_{(i,j) \in S, (i,k) \notin S} r_{ij} g(|\mathbf{h}_i - \mathbf{h}_j|, |\mathbf{h}_i - \mathbf{h}_k|)
\label{eq:mtrs}
\end{equation}
where $r_{ij} = |Y_i \cap Y_j| \ge 1$ denotes the number of labels $\mathbf{x}_i, \mathbf{x}_j$ share. 

\section{Derivation of the Unary Upper Bound}

\subsection{Multiclass Case (Detailed Derivation of Section 3.2)}

Considering that the class labels are evenly distributed, thus each label corresponds to $\frac{n}{C}$ data instances. Denote $\mathbf{c}_1, ..., \mathbf{c}_C$ as $C$ auxiliary vectors. We should prove that the Unary Upper Bound of the triplet loss defined in Eq. (\ref{eq:trf}) is
\begin{equation}
\begin{split}
\mathcal{L}_t \le (\frac{n}{C})^2 (C-1) \sum_{i=1}^n[ l_c(\mathbf{h}_i, y_i) +  2 |\mathbf{h}_i-\mathbf{c}_{y_i}| ] \\
l_c(\mathbf{h}_i, y_i) = \frac{1}{C-1} \sum_{l=1,l \neq y_i}^C g(|\mathbf{h}_i-\mathbf{c}_{y_i}|, |\mathbf{h}_i-\mathbf{c}_{l}|)
\end{split}
\label{eq:uub}
\end{equation}

In fact, we first of all address the triangle inequalities:
\begin{equation}
\begin{split}
&|\mathbf{h}_i-\mathbf{h}_j| \le |\mathbf{h}_i-\mathbf{c}_{y_i}| + |\mathbf{h}_j-\mathbf{c}_{y_j}|, \quad y_i=y_j \\
&|\mathbf{h}_i-\mathbf{h}_k| \ge |\mathbf{h}_i-\mathbf{c}_{y_k}| - |\mathbf{h}_k-\mathbf{c}_{y_k}|, \quad y_i \neq y_k
\end{split}
\label{eq:tri}
\end{equation}

Making use of the triangle inequalities above and the properties of function $g(\cdot, \cdot)$ shown in Eq. (\ref{eq:g}), we can arrive at the following inequality:
\begin{equation}
\begin{split}
g(|\mathbf{h}_i - \mathbf{h}_j|, |\mathbf{h}_i - \mathbf{h}_k|) &\le g(|\mathbf{h}_i-\mathbf{c}_{y_i}| + |\mathbf{h}_j-\mathbf{c}_{y_j}|, |\mathbf{h}_i - \mathbf{h}_k|) \\
&\le g(|\mathbf{h}_i-\mathbf{c}_{y_i}| + |\mathbf{h}_j-\mathbf{c}_{y_j}|, |\mathbf{h}_i-\mathbf{c}_{y_k}| - |\mathbf{h}_k-\mathbf{c}_{y_k}|) \\
&\le g(|\mathbf{h}_i-\mathbf{c}_{y_i}|, |\mathbf{h}_i-\mathbf{c}_{y_k}|-|\mathbf{h}_k-\mathbf{c}_{y_k}|) + |\mathbf{h}_j-\mathbf{c}_{y_j}| \\
&\le g(|\mathbf{h}_i-\mathbf{c}_{y_i}|, |\mathbf{h}_i-\mathbf{c}_{y_k}|) + (|\mathbf{h}_j-\mathbf{c}_{y_j}|+|\mathbf{h}_k-\mathbf{c}_{y_k}|) \\
& (y_i = y_j, y_i \neq y_k)
\end{split}
\label{eq:triest}
\end{equation}
where the first two inequalities in Eq. (6) holds according to the monotonous of $g(\cdot, \cdot)$ and the derivation of the last two inequalities are based on the Lipschitz continuity of $g(\cdot, \cdot)$.

It is clear that the data triplets can be generated by the following procedure:

\begin{itemize}
\item sample semantic label $s$;
\item sample semantic label $t \neq s$;
\item sample data instances $\mathbf{x}_i, \mathbf{x}_j$ such that $y_i = y_j = s$;
\item sample $\mathbf{x}_k$ such that $y_k = t$;
\end{itemize}

According to the sampling strategy above, Eq. (\ref{eq:trf}) can be reformulated as follows:
\begin{equation}
\begin{split}
\mathcal{L}_t &\le \sum_{(i,j) \in S, (i,k) \notin S} [ g(|\mathbf{h}_i-\mathbf{c}_{y_i}|, |\mathbf{h}_i-\mathbf{c}_{y_k}|) + (|\mathbf{h}_j-\mathbf{c}_{y_j}|+|\mathbf{h}_k-\mathbf{c}_{y_k}|) ] \\
&= \sum_{s=1}^C \sum_{t=1,t \neq s}^C \sum_{i: y_i = s} \sum_{j: y_j=s} \sum_{k: y_k = t} [ g(|\mathbf{h}_i-\mathbf{c}_s|, |\mathbf{h}_i-\mathbf{c}_t|) + (|\mathbf{h}_j-\mathbf{c}_s|+|\mathbf{h}_k-\mathbf{c}_t|) ]
\end{split}
\end{equation}

As $y_j, y_k$ is irrelevant with $\mathbf{h}_i$, we have $\sum_{j: y_j=s} \sum_{k: y_k = t}  g(|\mathbf{h}_i-\mathbf{c}_s|, |\mathbf{h}_i-\mathbf{c}_t|) = (\frac{n}{C})^2 g(|\mathbf{h}_i-\mathbf{c}_s|, |\mathbf{h}_i-\mathbf{c}_t|)$. And similar conclusions can be arrived such that $\sum_{i: y_i = s} \sum_{j: y_j=s} \sum_{k: y_k = t} |\mathbf{h}_j - \mathbf{c}_s| = (\frac{n}{C})^2 \sum_{j: y_j=s} |\mathbf{h}_j - \mathbf{c}_s|$ and $\sum_{i: y_i = s} \sum_{j: y_j=s} \sum_{k: y_k = t} |\mathbf{h}_k - \mathbf{c}_t| = (\frac{n}{C})^2 \sum_{k: y_k=t} |\mathbf{h}_k - \mathbf{c}_t|$. Thus the Unary Upper Bound of $\mathcal{L}_t$ is 
\begin{equation}
\begin{split}
\mathcal{L}_t &\le (\frac{n}{C})^2 [ \sum_{s=1}^C \sum_{t=1,t \neq s}^C \sum_{i: y_i = s} g(|\mathbf{h}_i-\mathbf{c}_s|, |\mathbf{h}_i-\mathbf{c}_t|) + \sum_{s=1}^C \sum_{t=1,t \neq s}^C  \sum_{j: y_j=s} |\mathbf{h}_j - \mathbf{c}_s| + \sum_{t=1}^C \sum_{s=1,s \neq t}^C \sum_{k: y_k=t} |\mathbf{h}_k - \mathbf{c}_t|] \\
&= (\frac{n}{C})^2 [\sum_{i=1}^n \sum_{t=1,t \neq y_i}^C g(|\mathbf{h}_i-\mathbf{c}_{y_i}|, |\mathbf{h}_i-\mathbf{c}_t|) + \sum_{j=1}^n \sum_{t=1,t \neq y_j}^C |\mathbf{h}_j - \mathbf{c}_{y_j}| + \sum_{k=1}^n \sum_{s=1,s \neq y_k}^C |\mathbf{h}_k - \mathbf{c}_{y_k}|] \\
&= (\frac{n}{C})^2 \sum_{i=1}^n [\sum_{t=1,t \neq y_i}^C g(|\mathbf{h}_i-\mathbf{c}_{y_i}|, |\mathbf{h}_i-\mathbf{c}_t|) + 2(C-1) |\mathbf{h}_i - \mathbf{c}_{y_i}|] \\
\end{split}
\end{equation}

The last equality can be arrived with $\sum_{t=1,t \neq y_i}^C |\mathbf{h}_i - \mathbf{c}_{y_i}| = (C-1) |\mathbf{h}_i - \mathbf{c}_{y_i}|$ as $t$ is irrelevant with $y_i$. Denote $l_c(\mathbf{h}_i, y_i) = \frac{1}{C-1} \sum_{t=1,t \neq y_i}^C g(|\mathbf{h}_i-\mathbf{c}_{y_i}|, |\mathbf{h}_i-\mathbf{c}_{t}|)$, then Unary Upper Bound of the triplet ranking loss is
\begin{equation}
\mathcal{L}_t \le (\frac{n}{C})^2 (C-1) \sum_{i=1}^n[ l_c(\mathbf{h}_i, y_i) +  2 |\mathbf{h}_i-\mathbf{c}_{y_i}|]
\end{equation}
which has the same formulation as Eq. (\ref{eq:uub}).

\subsection{Multilabel Case (Proof of Proposition 1)}

Similar with the triangle inequalities in Eq. (\ref{eq:tri}), the triangle inequalities for the mutlilabel dataset can be arrived such that:
\begin{equation}
\begin{split}
&|\mathbf{h}_i-\mathbf{h}_j| \le |\mathbf{h}_i-\mathbf{c}_{s}| + |\mathbf{h}_j-\mathbf{c}_{s}|, \quad s \in Y_i \cap Y_j \\
&|\mathbf{h}_i-\mathbf{h}_k| \ge |\mathbf{h}_i-\mathbf{c}_{t}| - |\mathbf{h}_k-\mathbf{c}_{t}|, \quad s \in Y_i, t \in Y_k, Y_i \cap Y_k = \emptyset
\end{split}
\end{equation}
thus we have
\begin{equation}
\begin{split}
g(|\mathbf{h}_i - \mathbf{h}_j|, |\mathbf{h}_i - \mathbf{h}_k|) \le g(|\mathbf{h}_i-\mathbf{c}_{s}|, |\mathbf{h}_i-\mathbf{c}_{t}|) + (|\mathbf{h}_j-\mathbf{c}_{s}|+|\mathbf{h}_k-\mathbf{c}_{t}|) \\
(s \in Y_i, s \in Y_j, t \in Y_k, Y_i \cap Y_k = \emptyset)
\end{split}
\end{equation}

Suppose the data triplets are generated by the following procedure:

\begin{itemize}
\item sample semantic label $s$;
\item sample semantic label $t \neq s$;
\item sample data instance $\mathbf{x}_i$ such that $s \in Y_i, t \notin Y_i$. $Y_i$ is defined in the Proposition 1;
\item sample $\mathbf{x}_j$ such that $s \in Y_j$;
\item sample $\mathbf{x}_k$ such that $t \in Y_k, Y_i \cap Y_k = \emptyset$.
\end{itemize}

Thus Eq. (\ref{eq:mtrs}) can be reformulated by the above sampling strategy:
\begin{equation}
\begin{split}
\mathcal{L}_{mt} & \le \sum_{(i,j) \in S, (i,k) \notin S} \sum_{s: s \in Y_i \cap Y_j} [g(|\mathbf{h}_i-\mathbf{c}_{s}|, |\mathbf{h}_i-\mathbf{c}_{t}|) + (|\mathbf{h}_j-\mathbf{c}_{s}|+|\mathbf{h}_k-\mathbf{c}_{t}|)] \qquad (t \notin Y_i, t \in Y_k) \\
&\le \sum_{s=1}^C \sum_{t=1, t\neq s}^C \sum_{\substack{i, j, k: s \in Y_i, s \in Y_j \\ t \in Y_k \\ Y_i \cap Y_k = \emptyset }} [ g(|\mathbf{h}_i - \mathbf{c}_s|, |\mathbf{h}_i - \mathbf{c}_t|) + |\mathbf{h}_j - \mathbf{c}_s| + |\mathbf{h}_k - \mathbf{c}_t|] \\
& = \sum_{s=1}^C \sum_{t=1, t\neq s}^C [ \sum_{j: s \in Y_j} \sum_{\substack{i: s \in Y_i, \\ t \notin Y_i}} \sum_{\substack{k: t \in Y_k, \\ Y_i \cap Y_k = \emptyset}} g(|\mathbf{h}_i - \mathbf{c}_s|, |\mathbf{h}_i - \mathbf{c}_t|) + \sum_{ \substack{i,k: s \in Y_i, \\ t\in Y_k, \\ Y_i \cap Y_k = \emptyset}} \sum_{j: s \in Y_j} |\mathbf{h}_j - \mathbf{c}_s| + \sum_{j: s \in Y_j} \sum_{ \substack{k: t \in Y_k \\ s \notin Y_k}} \sum_{\substack{i: s \in Y_i, \\ Y_i \cap Y_k = \emptyset }} |\mathbf{h}_k - \mathbf{c}_t| ]
\end{split}
\end{equation}

The second inequality holds in the fact that $\mathbf{x}_k$ is sampled for multiple times if $\mathbf{x}_k$ has multiple labels. Note that the probability $\mathrm{P}(s \in Y_i) = p$ satisfies for any $s = \{1,...,C \},i=\{1,...,n\}$, thus we can arrive at the following conclusions:
\begin{itemize}
\item for certain $s \in \{1,...,C\}$, the expected number of $j \in \{1,...,n\}$ such that $\{j: s \in Y_j\}$ is $pn$.
\item for certain $s \in \{1,...,C\}$ and $Y_i(i=1,...,n)$, the expected number of $k \in \{1,...,n\}$ such that $\{k: t \in Y_k, Y_i \cap Y_k = \emptyset \}$ is $p(1-p)^{|Y_i|} n$, where $|Y_i|$ is the number of elements of set $Y_i$.
\item for certain $s, t \in \{1,...,C\}$, the expected number of pairs $(i,k) \in \{1,...,n\} \times \{1,...,n\}$ such that $\{ (i,k): s \in Y_i, t \in Y_k, Y_i \cap Y_k = \emptyset\}$ is $p^2(1-p)^2(1-p^2)^{C-2} n^2$.
\end{itemize}

Then the upper bound of the expectation of $\mathcal{L}_{mt}$ is 
\begin{equation}
\begin{split}
\mathbb{E}[\mathcal{L}_{mt}] \le & \sum_{s=1}^C \sum_{t=1, t\neq s}^C [pn \sum_{\substack{i: s \in Y_i, \\ t \notin Y_i}} p(1-p)^{|Y_i|}n \cdot g(|\mathbf{h}_i - \mathbf{c}_s|, |\mathbf{h}_i - \mathbf{c}_t|) + p^2(1-p)^2(1-p^2)^{C-2} n^2 \sum_{j: s \in Y_j} |\mathbf{h}_j - \mathbf{c}_s| \\
&+ pn \sum_{\substack{k: t \in Y_k \\ s \notin Y_k}} p(1-p)^{|Y_k|}n |\mathbf{h}_k - \mathbf{c}_t| ] \\
=& p^2 n^2 \sum_{i=1}^n [ (1-p)^{|Y_i|} \sum_{s:s \in Y_i} \sum_{t:t \notin Y_i} g(|\mathbf{h}_i - \mathbf{c}_s|, |\mathbf{h}_i - \mathbf{c}_t|) \\
&+ ((C-1)(1-p)^2(1-p^2)^{C-2} + (C-|Y_i|)(1-p)^{|Y_i|}) \sum_{s: s \in Y_i} |\mathbf{h}_i - \mathbf{c}_s| ]
\end{split}
\end{equation}
where the last equality is derived by 
\begin{equation}
\begin{split}
\sum_{s=1}^C \sum_{t=1, t \neq s}^C \sum_{i: s\in Y_i, t \notin Y_i} g(|\mathbf{h}_i - \mathbf{c}_s|, |\mathbf{h}_i - \mathbf{c}_t|) = \sum_{i=1}^n \sum_{s \in Y_i} \sum_{t \notin Y_i} g(|\mathbf{h}_i - \mathbf{c}_s|, |\mathbf{h}_i - \mathbf{c}_t|) \\
\sum_{s=1}^C \sum_{t=1, t \neq s}^C \sum_{j: s\in Y_j} |\mathbf{h}_j - \mathbf{c}_s| = (C-1)\sum_{s=1}^C \sum_{j: s\in Y_j} |\mathbf{h}_j - \mathbf{c}_s| = (C-1) \sum_{j=1}^n \sum_{s \in Y_j} |\mathbf{h}_j - \mathbf{c}_s| \\
\sum_{s=1}^C \sum_{t=1, t \neq s}^C \sum_{k: t\in Y_k, s \notin Y_k} |\mathbf{h}_k - \mathbf{c}_t| = \sum_{t=1}^C \sum_{k: t\in Y_k} (C-|Y_k|)|\mathbf{h}_k - \mathbf{c}_t| = \sum_{k=1}^n \sum_{t \in Y_k} (C-|Y_k|) |\mathbf{h}_k - \mathbf{c}_t|
\end{split}
\end{equation}

Denote $l_{mc}(\mathbf{h}_i, Y_i) = \frac{1}{C-|Y_i|}\sum_{s \in Y_i} \sum_{t \notin Y_i} g(|\mathbf{h}_i - \mathbf{c}_s|, |\mathbf{h}_i - \mathbf{c}_t|), q(x) = \frac{C-x}{C-1}(1-p)^x, Q = (1-p)^2(1-p^2)^{C-2}$, the Unary Upper Bound of the  expectation value of $\mathcal{L}_{mt}$ is 
\begin{equation}
\mathbb{E}[\mathcal{L}_{mt}] \le (C-1)p^2 n^2 \sum_{i=1}^n [q(|Y_i|) l_{mc}(\mathbf{h}_i, Y_i) + (Q+q(|Y_i|)) \sum_{s \in Y_i} |\mathbf{h}_i - \mathbf{c}_s|]
\label{eq:mtuub}
\end{equation}
and it is clear that $l_{mc}(\mathbf{h}_i, Y_i)$ can be defined as the multilabel softmax loss in which
\begin{equation}
g(|\mathbf{h}_i - \mathbf{c}_s|, |\mathbf{h}_i - \mathbf{c}_t|) = - \log \frac{\exp (-|\mathbf{h}_i-\mathbf{c}_{s}|)}{\sum_{j=1}^C \exp(-|\mathbf{h}_i-\mathbf{c}_{j}|)} \quad i \in \{1,...,n\}, t \in \{1,...,C \}
\end{equation}
satisfies the condition in Eq. (\ref{eq:g}) and then
\begin{equation}
\begin{split}
l_{mc}(\mathbf{h}_i, Y_i) &= \frac{1}{C-|Y_i|} \sum_{s \in Y_i} \sum_{t \notin Y_i} [ -\log \frac{\exp (-|\mathbf{h}_i-\mathbf{c}_{s}|)}{\sum_{j=1}^C \exp(-|\mathbf{h}_i-\mathbf{c}_{j}|)} ]  \\
&= \frac{1}{C-|Y_i|} \sum_{s \in Y_i}  (C-|Y_i|) [ -\log \frac{\exp (-|\mathbf{h}_i-\mathbf{c}_{s}|)}{\sum_{j=1}^C \exp(-|\mathbf{h}_i-\mathbf{c}_{j}|)} ] \\
&= \sum_{s \in Y_i}  [ -\log \frac{\exp (-|\mathbf{h}_i-\mathbf{c}_{s}|)}{\sum_{j=1}^C \exp(-|\mathbf{h}_i-\mathbf{c}_{j}|)} ]
\end{split}
\end{equation}

Thus the proof of the proposition is completed.

\subsection{Discussions}

It is clear that the Unary Upper Bound defined above is established under the assumption that the data labels form a certain distribution. More specifically, the semantic labels should be evenly distributed in the multiclass case, and in the multilabel case, the number of labels should be almost the same and there are little relevance between labels. In practical applications, the labels in the dataset is unbalanced. In these cases, we can upsample certain data instances to make the labels balanced, which satisfies the assumptions shown above.

Moreover, motivated by Eq. (\ref{eq:mtuub}), similar as Section 3.3, we can arrive at a more general form of the Unary Upper Bound such that
\begin{equation}
\begin{split}
\mathbb{E}[\mathcal{L}_{mt}] \le M_{mt} \mathcal{L}_{mu} \\
\mathcal{L}_{mu} = \sum_{i=1}^n [s(|Y_i|) l_{mc}(\mathbf{h}_i, Y_i) + u(|Y_i|) \sum_{s \in Y_i} |\mathbf{h}_i - \mathbf{c}_s|]
\end{split}
\label{eq:mtuub2}
\end{equation}
where $s(x), u(x)$ are non-negative. Although we just arrive at the bound of the expected triplet loss value, this form of Unary Upper Bound defined in Eq. (\ref{eq:mtuub2}) is able to be adopted in many practical applications. In Section 4, we just incorporate $\mathcal{L}_{mu}$ with $s(x) = 1/x, u(x) = \mathrm{constant}$ in the proposed SCDH algorithm and achieves the state-of-the-art hashing results on the multilabel dataset.

\end{appendices}

\end{document}